\RequirePackage{fix-cm}
\documentclass[final,smallextended]{svjour3}

\smartqed

\usepackage{graphicx}

 \usepackage{subfig}
 \usepackage{epsfig}
 \usepackage[utf8]{inputenc}
 \usepackage{amsmath,amssymb,amsfonts,nomencl, mathtools} 
 \usepackage{booktabs}
 \usepackage{algorithmicx}
 \usepackage{amsmath}

 \usepackage{amstext,amsmath,amssymb}
 \usepackage{algorithm}
 \usepackage{hyperref}
 \usepackage{algpseudocode}
 \usepackage{xcolor}
 \usepackage{natbib}
 \usepackage{multirow}
 \usepackage{color,xcolor}

 \usepackage{tikz}
\usepackage{pgfplots}

\usepackage{appendix}

\graphicspath{{figs/}}

\makeatletter
\newcommand*\bigcdot{\mathpalette\bigcdot@{.5}}
\newcommand*\bigcdot@[2]{\mathbin{\vcenter{\hbox{\scalebox{#2}{$\m@th#1\bullet$}}}}}

\usepackage[markup=none,commentmarkup=footnote]{changes}
\usepackage{color}
\definechangesauthor[color=red]{YS}

\definecolor{marine}{RGB}{0,32,96}
\definecolor{navy}{RGB}{0,0,128}
\definecolor{maroon}{RGB}{128,0,0}
\definecolor{olivegreen}{RGB}{85,107,47}
\definecolor{gray}{RGB}{102,102,102}
\definecolor{green}{RGB}{131,198,210}
\definecolor{blue}{rgb}{0, 0.4470, 0.7410}
\definecolor{skyblue}{rgb}{0.3010, 0.7450, 0.9330}
\definechangesauthor[color=maroon]{AE}
\definechangesauthor[color=blue]{XL}

\begin{document}


\title{Global-Supervised Contrastive Loss and View-Aware-Based Post-Processing for Vehicle Re-Identification}

\author{Zhijun Hu \and Yong Xu \and Jie Wen \and Xianjing Cheng \and Zaijun Zhang \and Lilei Sun \and Yaowei Wang
}

\institute{Zhijun Hu (Corresponding Author) \at
         School of Mathematics and Statistics, Guangxi Normal University, Guilin, 541004, China. \\ College of Computer Science and Technology, Guizhou University, Guiyang, 550025,China. \\
        \email{huzhijun@mailbox.gxnu.edu.cn; sunlileisun@163.com}           
        \and
        Yong Xu \and Jie Wen \at
        Shenzhen Key Laboratory of Visual Object Detection and Recognition, Shenzhen, China \\
        \email{laterfall@hit.edu.cn; jiewen\_pr@126.com} 
        \and
        Xianjing Cheng \and Lilei Sun \at
        College of Computer Science and Technology, Guizhou University, Guiyang, 550025,China. \\
        \email{chengxianjing2014@126.com; sunlileisun@163.com}
        \and
        Zaijun Zhang \at
        Key Laboratory of Complex Systems and Intelligent Computing and School of Mathematics and Statistics, Qiannan Normal University for Nationalities, Duyun 558000, China. \\
        \email{zjzhang1987@outlook.com}
        \and 
        Yaowei Wang \at
        Peng Cheng Laboratory, Shenzhen, 518055, China.
        \email{wangyw@pcl.ac.cn}
}

\date{Received: date / Accepted: date}



\maketitle

\section*{Abstract}
	In this paper, we propose a Global-Supervised Contrastive loss ($\mathcal{L}_{GSupCon}$) and a view-aware-based post-processing (VABPP) method for the field of vehicle re-identification. The traditional supervised contrastive loss ($\mathcal{L}_{SupCon}$) calculates the distances of features within the batch, so it has the local attribute. While the proposed $\mathcal{L}_{GSupCon}$ has new properties and has good global attributes, the positive and negative features of each anchor in the training process come from the entire training set. The proposed VABPP method is the first time that the view-aware-based method is used as a post-processing method in the field of vehicle re-identification. The advantages of VABPP are that, first, it is only used during testing and does not affect the training process. Second, as a post-processing method, it can be easily integrated into other trained re-id models. We directly apply the view-pair distance scaling coefficient matrix calculated by the model trained in this paper to another trained re-id model, and the VABPP method greatly improves its performance, which verifies the feasibility of the VABPP method.

\section{Introduction}
\label{intro}
Due to the vigorous development of deep learning, deep learning methods have penetrated into many research fields \cite{ref3,ref2,ref4}, so does as the field of vehicle re-identification. Vehicle re-identification aims to find the same vehicle as the query vehicle in non overlapping cameras \cite{ref5}. Although the license plate can be used as the unique identification of the vehicle, it is usually difficult to capture the license plate correctly due to factors such as occlusion, illumination and camera distance. Vehicle re-identification has attracted more and more attention because it only uses vehicle appearance to identify vehicles.
\begin{figure}[!t]
	\centering
	\includegraphics[width=.75\textwidth]{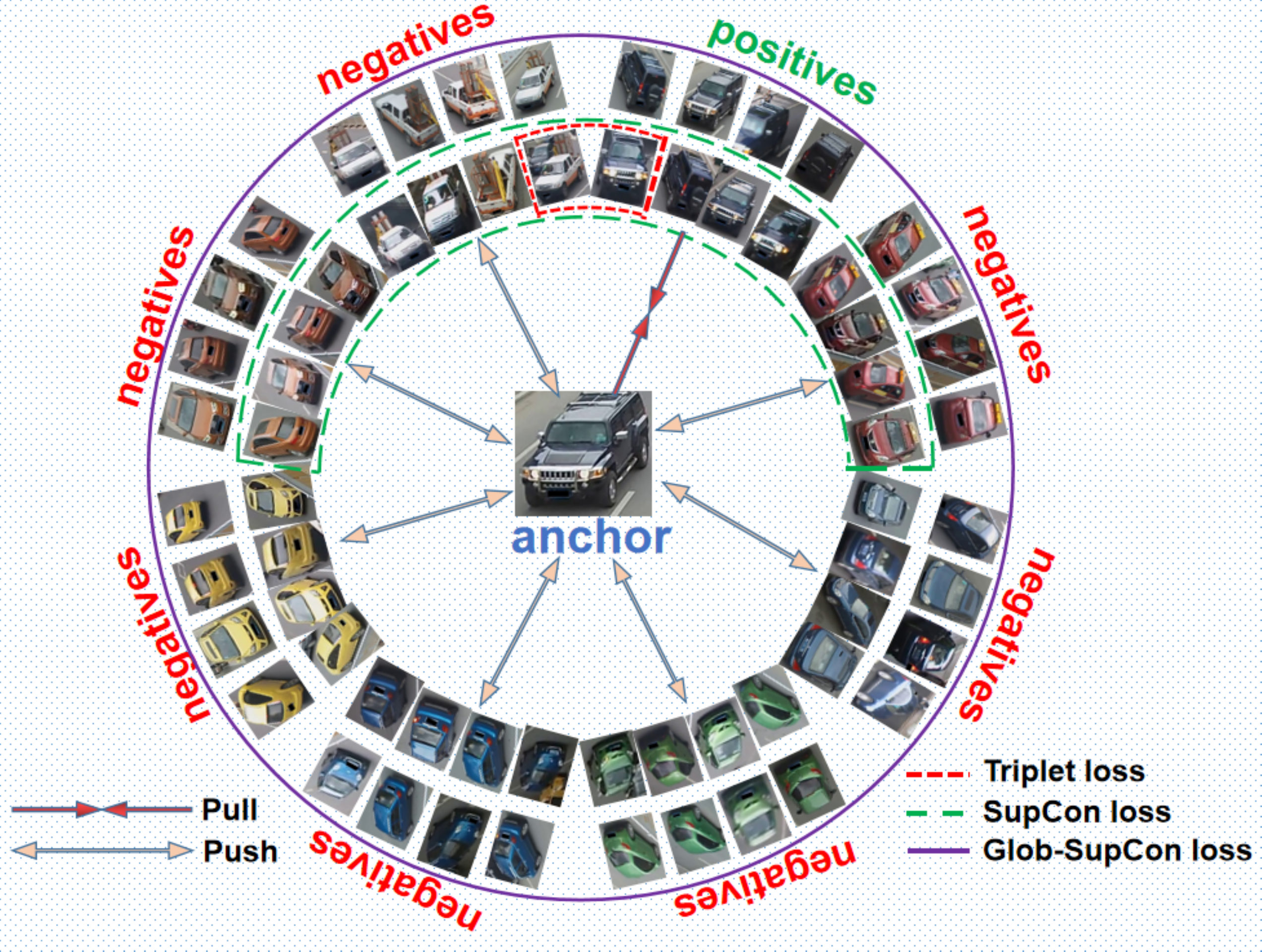}%
	\caption{Triplet loss, $\mathcal{L}_{SupCon}$ and $\mathcal{L}_{GSupCon}$. The green dotted line represents samples in a batch, and the purple line represents the entire training set. Triplet loss only uses a positive and a negative in the batch. The $\mathcal{L}_{SupCon}$ considers all positives and all negatives in the batch. The $\mathcal{L}_{GSupCon}$ considers all positives and all negatives in the entire training set.}
	\label{tri_sup_glob}
\end{figure}

The two biggest challenges faced by vehicle re-identification are, first, how to design a metric learning method to extract more discriminative vehicle features. Second, the visual differences of vehicle appearance are very large. The same vehicle may look very different from different views, while different vehicles of the same model and color produced by the same manufacturer may look very similar from the same view. For the first challenge, due to the good performance, triplet loss \cite{ref6,ref7} has been used in the field of vehicle re-identification for a long time \cite{ref8,ref9}. 
However, for triplet loss, an anchor only has one positive sample and one negative sample, which limits the performance of triplet loss. The advantage of $\mathcal{L}_{SupCon}$ \cite{ref10} is that for an anchor, it makes use of all positive samples and all negative samples in the batch, which increases the stability of training \cite{ref10}. Since references \cite{ref10,ref11,ref12} claimed that increasing the number of negative IDs can improve performance, we hope to continue to increase the number of negative IDs on the basis of $\mathcal{L}_{SupCon}$. We proposed a loss function named Global-SupCon Loss ($\mathcal{L}_{GSupCon}$), for an anchor, \textbf{the proposed $\mathcal{L}_{GSupCon}$ extends the positives and negatives to the entire training set} (Fig. \ref{tri_sup_glob}).

For the second challenge, there have been many researches on view-aware-based method \cite{ref13,ref14,ref15}, which were all conducted by adding orientation suppression to the model parameters in the training process to make the model parameters learned the view-aware knowledge to improve the model performance. Such methods are difficult to be integrated into other methods. This paper proposes a view-aware-based post-processing method (VABPP). As far as we know, this is the first time that view-aware method is proposed as a post-processing method. \textbf{The proposed VABPP aims to improve the experimental accuracy during testing by scaling the distance between the features of the images in the gallery set and the feature of the query image, and can be easily integrated into other methods}.

\begin{figure}[!t]
	\centering
	\includegraphics[width=.85\textwidth]{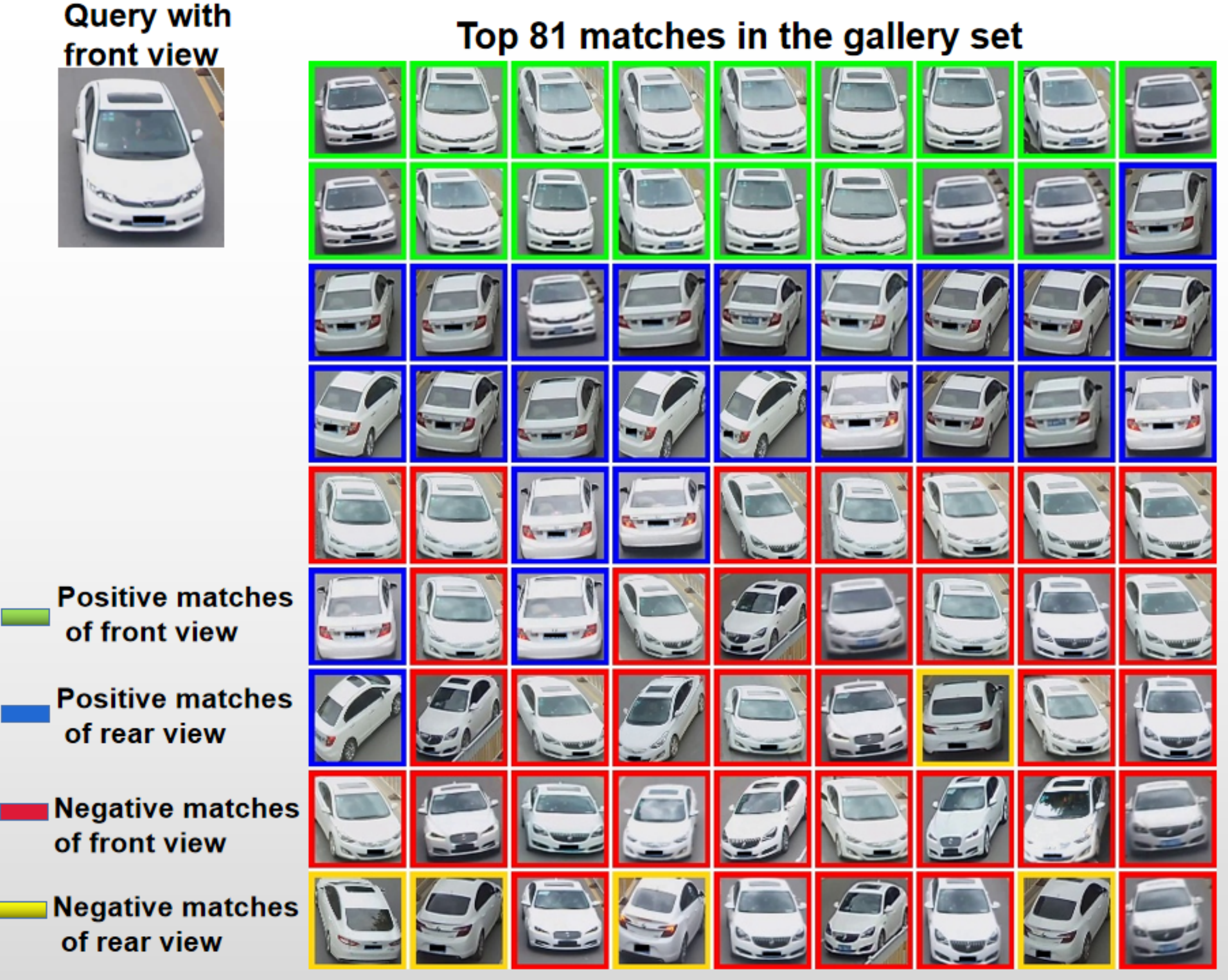}%
	\caption{A matching result of the baseline model. The matching result is sorted in ascending order of the Euclidean distance between the features of the query image and the features of the images in the gallery set.}
	\label{motivation2}
\end{figure}

Fig. \ref{motivation2} shows the motivation of the VABPP method. Based on our baseline, Fig. \ref{motivation2} shows the matching results of a query image q matching in the gallery set. If all the rear view vehicles (blue boxes) are moved forward by the distance equal to the length of 19 images, so that the green boxes and the blue boxes are crosswise ranked, and we can find the last blue image is moved before the first red image, that is, all positive matches are moved before all negative matches, which significantly improves the matching accuracy. Note that for the sake of intuition, the above introduction is based on translation, but scaling is used later of this paper.

To sum up, the main contributions of this paper are as follows:

(1)We design a Global-Supervised Contrastive loss, `all positive and negative features do not have the property of gradient, and the positive and negative features of each anchor are from the entire training set.

(2) We design a view-aware-based post-processing method to improve the test accuracy by stretching (or shrinking) the distances between the features of the images with the same view in the gallery set and the feature of the query image in a whole with a distance scaling coefficient. The distances between the features of the images of the same view and the anchor feature have the same scaling coefficient, while the distances between the images with different views and the anchor have different scaling coefficients. This method can be easily integrated into other methods without affecting the training process.

(3) The proposed method is verified on the three data sets widely used in vehicle re-identification, and our method achieves the state-of-the-art level.

The rest of this paper is arranged as follows. The second part reviews some works related to this paper. The third part introduces the main idea of the proposed method. The fourth part expounds the proposed method in detail. The fifth part analyzes the reasons why the proposed method is effective by experiments and compares it with the state-of-the-art methods. The sixth part summarizes the full paper.

\section{Related work}
\subsection{Vehicle re-identification based on metric learning}
Metric learning method is usually packaged into the form of loss function, so that it can be used directly in the program. The loss functions commonly used in the field of vehicle re-identification include contrastive loss and triplet loss. The advantage of contrastive loss is that it can increase the distances between features of different classes while decrease the distances between features of the same class. Shen et al. \cite{ref17} utilized LSTM network to memorize path, and utilized Siamese neural network to regularize similarity scores for robust re-identification performance. Zhu et al. \cite{ref18} proposed using Siamese neural network structure to simultaneously extract the deep features of input vehicle image pairs under the supervision of joint identification and verification. Zhu et al. \cite{ref19} proposed a densely connected convolutional neural network for vehicle re-identification, which adopted the structure of Siamese neural network and included two deep feature learning branches with shared parameters, which effectively improved the feature learning ability. The disadvantage of contrastive loss is that, for an anchor, a positive pair and a negative pair are randomly selected for the batch, which weakens the training mechanism and the training speed is particularly slow. The advantage of triplet loss is that it has a hard mining mechanism  which speeds up the training speed. Bai et al. \cite{ref20} proposed that in triplet network learning, by adding an intermediate representation "group" between the sample and each vehicle to model the intra-class variance, divided the samples of each vehicle into several groups, and establish multi granularity triplet samples between different vehicles and different groups in the same vehicle to learn fine-grained features. Kumar et al. \cite{ref21} used triplet loss to solve the problem of vehicle re-identification, and introduced the formal evaluation of triplet sampling variants (batch samples) into the re-identification task. Lou et al. \cite{ref22} proposed coupling re-id model to feature distance adversarial network, and designed a new feature distance adversarial scheme to generate hard negative samples online in feature space. Ghosh et al. \cite{ref23} introduced relationship preserving triplet mining (RPTM), which is a triplet mining scheme guided by feature matching to ensure that triples respect the natural subgroups in the object ID, and used this triplet mining mechanism to establish vehicle pose estimation to form a triplet cost function. The disadvantage of triplet loss is that for each anchor, there is only one positive sample and one negative sample, while ignoring other positive and negative samples. For the supervised contrastive loss, for each anchor, all the positive and all the negative samples in the batch were used. Huynh et al. \cite{ref24} formed a strong baseline by applying the supervised contrastive loss and network with multi head method for the field of vehicle re-identification.

\subsection{vehicle re-identification based on view-aware methods}
Compared with person re-identification, the biggest challenge of vehicle re-identification is that the appearance of vehicles change greatly with the change of viewpoint. In order to overcome this difficulty, scholars have proposed many view-aware based methods. The keypoint based attention model can judge the orientation information according to the keypoint information. Wang et al. \cite{ref25}, Khorramshahi et al. \cite{ref13} and Zheng et al. \cite{ref14} extracted local region features in different orientations based on the location of keypoints, and combines the global features to form orientation invariant features. However, the above methods required expensive keypoint annotations and were difficult to implement when the two vehicle images don’t have any common visible area. So an effective approach is to employ a GAN network to generate vehicle images in invisible view. Zhou et al. \cite{ref26,ref27} and Pan et al. \cite{ref28} utilized convolutional neural networks (CNN) and long short-term memory (LSTM) to learn the transition between different vehicle viewpoints, and can infer the vehicle features containing all view information from one view. Zhou et al. \cite{ref29} and Lou et al. \cite{ref30} designed cross-view generative adversarial networks to efficiently infer cross-view images, combining the features of the original images with the features of the generated cross-view images to learn vehicle re-identification distance metric. However, at present, the generative adversarial network works have poor effect on objects with large parallax changes such as vehicles. The view-aware based embedding network can use the view label information to narrow the distance between features of images of the same vehicle with different views, while pushing the distance between features of images of different vehicles with the same view. Wang et al. \cite{ref32}, Chu et al. \cite{ref15} and sun et al. \cite{ref16} designed different branches to extract the features of different viewpoints (or spaces) respectively, and use the triplet loss to narrow the distance between the features of images of the same vehicle, and simultaneously increase the distance between different vehicles. Teng et al. \cite{ref33} designed a multi-view and multi-branch network, each branch learned the features of each viewpoint, and combined a spatial attention model to enhance the discriminativity of features. Zhu et al. \cite{ref34} first extracted the vehicle features and orientation features of the vehicle image, then the distance between orientation features was subtracted from the distance between vehicle features, so as to reduce the difference caused by orientation. Chen et al. \cite{ref35} proposed a special Semantics-guided Part Attention Network (SPAN) to robustly predict the part attention masks for different views of vehicles, so as to achieve the purpose of adaptive view aware. Li et al. \cite{ref36} generated potential view labels through clustering and considers view information to improve vehicle re-identification performance. Jin et al. \cite{ref37} first introduced several potential view clusters for a vehicle to simulate potential multi-view information. Each view cluster had a learnable center.

\section{The main idea of the proposed method}
In this section, we will introduce the idea of $\mathcal{L}_{GSupCon}$ and VABPP.

\subsection{The main idea of $\mathcal{L}_{GSupCon}$}

The proposed $\mathcal{L}_{GSupCon}$ has different properties from the traditional supervised contrastive loss ($\mathcal{L}_{SupCon}$). $\mathcal{L}_{SupCon}$ is essentially a local function related to batch samples. The positives and negatives of an anchor only come from the batch, so in this paper, the $\mathcal{L}_{SupCon}$ is called local supervised contrastive loss, and can be denoted as $\mathcal{L}_{LSupCon}$. While in our proposed $\mathcal{L}_{GSupCon}$, by changing the gradient properties of the positive and negative features, the positives and negatives of each anchor can come from the entire training set, so that the entire training set contributes to the training of the anchor, thus greatly enhancing the global property of $\mathcal{L}_{GSupCon}$, which makes the training process easier to converge to a global stable point. In $\mathcal{L}_{LSupCon}$, the parameter optimization process makes each anchor feature move to the local optimal point in the batch. While in $\mathcal{L}_{GSupCon}$, the parameter optimization process makes each anchor feature move to the global optimal point in the entire training set, the anchor is close to the positive features of the entire training set and is also far from the negative features of the entire training set.

Reference \cite{ref10,ref11,ref12} has explained that increasing the negative IDs can improve the performance. If we want to increase the negative images in $\mathcal{L}_{LSupCon}$, we can only achieve this by increasing the batch size. During forward and backward of training, each layer of the deep model will store gradients generated related to the newly added batch samples, which will occupy a lot of memory and make the program unable to run. Masters et al. \cite{ref38} and Ge et al. \cite{ref39} have explained that, a smaller batch size can make the model better jump out of the local optimal solution during training, so increasing batch size is not conducive to model convergence. But if we increase the number of positives and negatives in the proposed $\mathcal{L}_{GSupCon}$, by doing this, we don't increase the batch size, and the program will only increase a small amount of memory to store these increased features, these memory increments have no relationship with the depth of the model, because we remove all the gradient attributes of all positive features and all negative features,  during training process, the program will not allocate memory to store the gradients corresponding to these newly added features.

\subsection{The main idea of VABPP}
\label{sec32}
The ultimate purpose of the VABPP method is to reduce the matching difference caused by different views during the testing process, so in the following sections, all the introductions about this method, except the calculation of the view-pair distance scaling coefficient matrix  $\Delta^{V\times V}$ in section \ref{sec43} is under the training set, the rest are all based on the test set. The key step of the VABPP method is to find out the center of the distances between features of all positive images with the same view and the feature of the query image, for this query image, there is a maximum of $V$ such centers, where $V$ is the number of views of the dataset, and then scale all such centers to the same position with corresponding scaling coefficients, then find out all images in the gallery set with the view corresponding to each center, and scale the distances between the features of these images and the feature of the query image with the corresponding scaling coefficient.

In Fig. \ref{motivation2}, we can see in the matching results, first, all positive images with the same view have a high probability of being compactly ranked together, second, all positive images with different views are probably not cross-ranked, and thirdly, the top-ranked false matches and the top-ranked correct matches have a high probability of having the same view as the query image (such as the front view in Fig. \ref{motivation2}), the reason is that images with this view are easier to match for query images

The VABPP method hopes that in all correct matches, images with different views are cross-ranked. If the number of images of different views of a vehicle is the same, ideally, in the same interval, it is hoped that the same number of images of each view of this vehicle are included. 

\begin{figure}[!t]
	\centering
	\includegraphics[width=.85\textwidth]{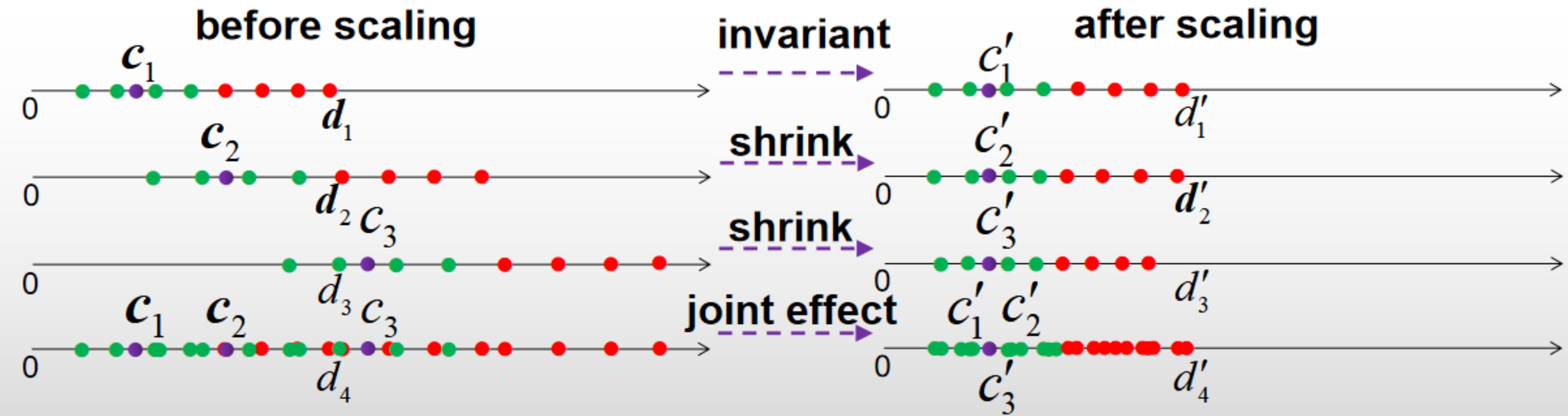}%
	\caption{The effect of VABPP. The purpose of VABPP is to make $c_1$, $c_2$ and $c_3$ on the left, after scaling, become $c^{\prime}_1$, $c^{\prime}_2$ and $c^{\prime}_3$ on the right, and $c^{\prime}_1=$$c^{\prime}_2=$$c^{\prime}_3=c_1$, that is, all the distances in the above three lines, are scaled according to the formula $d^{\prime}_i=d_i\frac{c_1}{c_i}(i=1,2,3)$. Where $c_i$ and $c^{\prime}_i (i=1,2,3)$ represents the average distances of the four green points in the $i$-th row on the left and the right, respectively, where the green points and red points represent the distances between the positive features and the query feature, and distances between the negative features and the query feature, respectively. The query image has view $v_1$, The first row, the second row and the third row respectively represent the distance between the features all the images in the gallery set with views $v_1, v_2$ and $v_3$ ($v_1, v_2$ and $v_3$ are unequal to each other) and the feature of query image and the fourth row represents the schematic diagram of the distances from the first row, the second row, and the third row put together. Since the images represented in the first row and the query image have the same view, the distances of the first row do not change. The left and right represent the situation before and after applying the VABPP method, respectively. It can be seen that the fourth row on the right (compared to the left) has clearly separated the positives and negatives.}
	\label{movement3}
\end{figure}

In order to achieve this purpose, we need to scale the distances between the features of the images with different views and the feature of query image with different scaling coefficients (i.e., multiply the distance by a scaling coefficient, for example, $d^{\prime}=0.5d$ means to change the distance to 0.5 times of the original distance, 0.5 is the scaling coefficient), and the scaling coefficients of the distances between the features of the images with the same view and the feature of the query image is the same (that is, the rankings between all images with the same view are unchanged), if all the positive images with a certain view are ranked relatively low (such as the rear view images in Fig. \ref{motivation2}), then the rankings of all images with this view are moved forward as a whole (Fig. \ref{movement3})). It is reflected in Fig. \ref{motivation2} that the distances between the features of the images with rear views and the feature of the query image is shrunk, so that all images with rear view are moved forward. It is easy to understand that the more difficult to match the images of a certain view, the lower the rankings (such as the rear view images in Fig. \ref{motivation2}), the closer the required scaling coefficient to 0 to moved these images to  specified position (scaling coefficient is usually a positive value greater than 0 and less than 1 ).And we stipulate that the distances between the features of the images with the same view as the query image and the feature of the query image is not scaled, that is, the scaling coefficient is 1.

The advantage of this is that the scaling coefficient of the distance between the feature of a negative image which is easy to be wrong matched and the feature of the query image is usually larger (closer to 1), while the scaling coefficient of the distance between the feature of a positive image that is difficult to match correctly and the feature of the query image is smaller (closer to 0), so after applying the VABPP method, the rankings of the positives that are difficult to match correctly move more forward than the rankings of the negative images that are more likely to be incorrectly matched, which improves the test performance (Fig. \ref{movement3}). ).

Traditional view-aware-based vehicle re-identification methods are used in the training process, so that these methods are difficult integrated into other non-view-aware-based methods. The advantages of the VABPP are that, first, it does not affect the training process, and second, the VABPP can be easily integrated into other methods.

\section{Proposed method}
\begin{figure*}[!t]
	\centering
	\includegraphics[width=\textwidth]{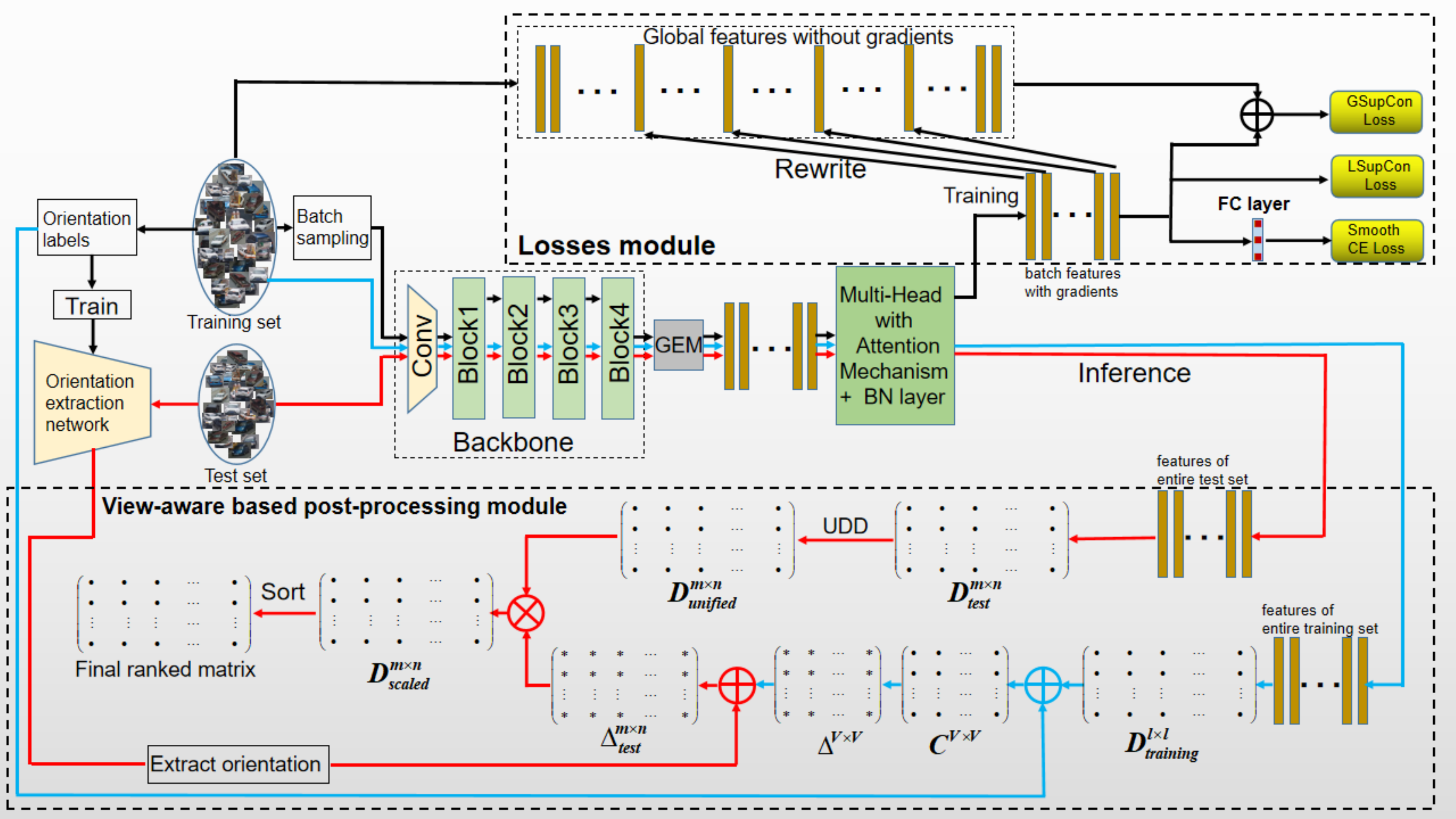}%
	\caption{Our pipeline. Our pipeline includes the losses module and the view-aware-based post-processing module. The losses module shows the pipeline of our proposed $\mathcal{L}_{GSupCon}$ method. The view-aware-based post-processing module shows the pipeline of our proposed VABPP method. The black arrow in the figure represents the training of the re-id network, and the blue arrow represents the calculation process of the view-pair distance scaling coefficient matrix $\Delta^{V\times V}$. The red arrow represents the application of VABPP method in the test set. $l,m$ and $n$ in the figure represent the number of images in the training set, query set and gallery set respectively. '.' represents the distance and '*' represents the view-pair distance scaling coefficient.}
	\label{network}
\end{figure*}

Below we introduce the details of the method proposed in this paper. We first use the proposed $\mathcal{L}_{GSupCon}$ to train the re-id model, and then apply the VABPP on the trained model. Therefore, we first introduce $\mathcal{L}_{GSupCon}$, then introduce the details of VABPP, and then introduce the steps of integrating the VABPP method into other arbitrary trained model. Fig. \ref{network} is a framework of the entire proposed method.

\subsection{Global Supervised Contrastive loss}
\label{sec_lgsupcon}
The formula of $\mathcal{L}_{LSupCon}$ is as follows:
\begin{equation}
	\label{LL}
	\mathcal{L}_{SupCon}=\sum_{i\in I}\frac{-1}{\lvert P(i) \rvert}\sum_{p\in P(i) \subset I}log\frac{exp(f_if_p/\tau)}{\sum_{a\in A(i)\subset I}exp(f_if_a/\tau)}
\end{equation}

\noindent Where $I$ is the set of all images in the batch, $A(i)=I\backslash\{i\}$, $P(i)\equiv\{p\in A(i):y_p=y_i\}$ is the set of all images in the batch that are different from image i and have the same vehicle ID as image $i$. $y_p$ and $y_i$ represent the ground truth labels of images $p$ and $i$, respectively. $\lvert P(i)\rvert$ is the cardinality. $\tau$ is a scalar temperature parameter. $f_i$ is an anchor feature, $f_p$ is a positive feature and $f_a$ is a feature of any image in the batch that is different from $f_i$.

The formula of our proposed $\mathcal{L}_{GSupCon}$ is as follows:

\begin{equation}
	\label{LG}
	\mathcal{L}_{GSupCon}=\sum_{i\in I}\frac{-1}{\lvert \widetilde{P}(i) \rvert}\sum_{p\in \widetilde{P}(i)}log\frac{exp(f_i\widetilde{f}_p/\tau)}{\sum_{a\in T}exp(f_i\widetilde{f}_a/\tau)}
\end{equation}

\noindent where $T$ is the training set, and $\widetilde{P}(i)\equiv\{p\in T:y_p=y_i\}$ is the set of all positive images of $i$ in the training set. $f_i\in I$ is the anchor feature which has gradient attribute (i.e. the $\frac{\partial\mathcal{L}_{GSupCon}}{\partial f_i}$ is generally not zero). $\widetilde{f}_p$ is a positive feature without gradient attribute (i.e. $\frac{\partial\mathcal{L}_{GSupCon}}{\partial\widetilde{f}_i}$ is always zero), $\widetilde{f}_a$ is feature of $a$ without gradient attribute, where $a\in T$, in the training process, we use a global dictionary to store all global features , and during each iteration, we use the features in the batch to update the corresponding features in the global dictionary.

\begin{figure}[!t]
	\centering
	\includegraphics[width=.85\textwidth]{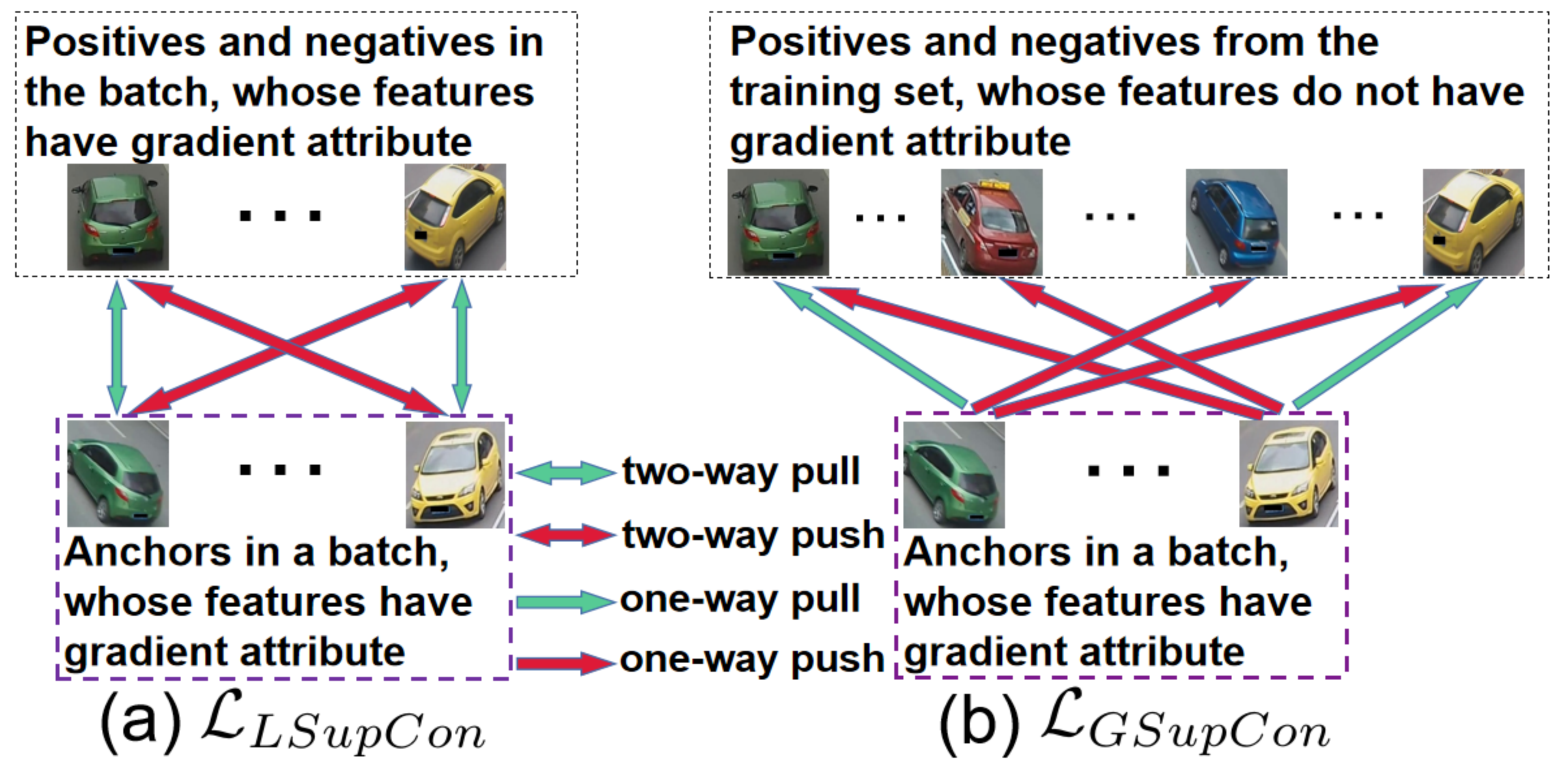}%
	\caption{The difference between $\mathcal{L}_{LSupCon}$ and $\mathcal{L}_{GSupCon}$. In $\mathcal{L}_{LSupCon}$, positive features and negative features have derivative property, while in $\mathcal{L}_{GSupCon}$, they do not have derivative property.  $\mathcal{L}_{LSupCon}$ is a local-to-local loss with two-way movement, while $\mathcal{L}_{GSupCon}$ is a local-to-global loss with one-way movement.}
	\label{glob_local_supcon_move}
\end{figure}

The parameter optimization process of $\mathcal{L}_{LSupCon}$ has the property of Local-to-Local two-way movement between features. That is, the parameter optimization process will make the anchor feature close to/away from the positive/negative features, but meanwhile, the positive/negative features are also close to/away from the anchor feature, and all these movements are conducted within batch features. The parameter optimization process of $\mathcal{L}_{GSupCon}$ has the property of Local-to-Global one-way movement of only the feature in the batch (anchor) close to/away from the positive/negative features in the entire training set (Fig. \ref{glob_local_supcon_move}). This can be observed by the gradients of  $\mathcal{L}_{LSupCon}$ and  $\mathcal{L}_{GSupCon}$ with respect to $f_i$ in equations \ref{LL} and \ref{LG}, respectively (see appendix A for the solutions process of these two gradients):

\begin{equation}
	\label{diff_lg_zi}
	\frac{\partial\mathcal{L}_{GSupCon}}{\partial f_i}=\frac{1}{\tau}\{\sum_{p\in \widetilde{P}(i)}\widetilde{f}_p[\widetilde{\Gamma}_{ip}-\frac{1}{\lvert\widetilde{P}(i)\rvert}]+\sum_{n\in\widetilde{N}(i)}\widetilde{f}_n\widetilde{\Gamma}_{in}\}
\end{equation}
\begin{equation}
	\begin{aligned}
		\label{diff_ll_zi}
		\frac{\partial\mathcal{L}_{LSupCon}}{\partial f_i}=\frac{1}{\tau}\{\sum_{p\in P(i)}f_p[\Gamma_{ip}-\frac{1}{\lvert P(i)\rvert}]+\sum_{n\in N(i)}f_n\Gamma_{in}\}\\ + \frac{1}{\tau}\{\sum_{p\in P(i)}f_p[\Gamma_{pi}-\frac{1}{\lvert P(p)\rvert}]+\sum_{n\in N(i)}f_n\Gamma_{ni}\} 
	\end{aligned}
\end{equation}

\noindent Where $\widetilde{\Gamma}_{xy}=\frac{exp(f_x\widetilde{f}_y/\tau)}{\sum_{a\in T}exp(f_x\widetilde{f}_a/\tau)}$ and $\Gamma_{xy}=\frac{exp(f_xf_y/\tau)}{\sum_{a\in A(x)}exp(f_xf_a/\tau)}$.

The training should converge, so we have $\frac{\partial\mathcal{L}_{GSupCon}}{\partial f_i}\to 0$ and $\frac{\partial\mathcal{L}_{LSupCon}}{\partial f_i}\to 0$, finally we have each term of equations \ref{diff_lg_zi} and \ref{diff_ll_zi} should tend to 0. In equation \ref{diff_lg_zi}, $\widetilde{\Gamma}_{ip}-\frac{1}{\lvert \widetilde{P}(i)\rvert} \to 0$ and $\widetilde{\Gamma}_{in}\to 0$ show that in the $\mathcal{L}_{GSupCon}$, the anchor feature is close to all the positive features of the entire training set and away from all the negative features of the entire training set.
In equation \ref{diff_ll_zi}, $\Gamma_{ip}-\frac{1}{\lvert P(i)\rvert} \to 0$ and $\Gamma_{in} \to 0$ show that in the  $\mathcal{L}_{LSupCon}$, the anchor feature is close to all the positive features in the batch and is far away from all the negative features in the batch. Meanwhile, $\Gamma_{pi}-\frac{1}{\lvert P(p)\rvert} \to 0$ and $\Gamma_{ni} \to 0$ show that all the positive features in the batch are close to the anchor feature and all the negative features in the batch are far away from the anchor feature.

\textbf{\textit{$\mathcal{L}_{GSupCon}$ VS $\mathcal{L}_{LSupCon}$}} (1) The purpose of $\mathcal{L}_{GSupCon}$ is to improve the performance by increasing the number of negative IDs, so generally speaking, the larger the training set, the better the performance.
(2) $\mathcal{L}_{LSupCon}$ has the property of Local-to-Local two-way movement and $\mathcal{L}_{GSupCon}$ has the property of Local-to-Global one-way movement. If $\mathcal{L}_{LSupCon}$ and $\mathcal{L}_{GSupCon}$ are combined together to train the model, this Local-to-Local two-way movement mode and Local-to-Global one-way movement mode will promote each other when training with a smaller training set, while $\mathcal{L}_{LSupCon}$ will restrict the performance of $\mathcal{L}_{GSupCon}$ when training with a larger training set.

Since the proposed $\mathcal{L}_{GSupCon}$ is better for larger training set, we use the following weighted loss to train the re-id model for the dataset with a larger training set:
\begin{equation}
	\mathcal{L}=\lambda_{ID}\mathcal{L}_{ID}+\lambda_{Metric}\mathcal{L}_{GSupCon}
	\label{small_L}
\end{equation}
The same as our baseline, we use the label smooth cross entropy loss as our ID loss, i.e. $\mathcal{L}_{ID}$ in formula \ref{small_L} is ID loss. And in formula \ref{small_L}, the same as our baseline, $\lambda_{ID}$ and $\lambda_{Metric}$ are calculated by Momentum Adaptive Loss Weight method \cite{ref24,ref40}. While for smaller training set, we use the following weighted loss to train the re-id model:

\begin{equation}
	\mathcal{L}=\lambda_{ID}\mathcal{L}_{ID}+\lambda_{Metric}\mathcal{L}_{GSupCon}+\lambda_{Metric}\mathcal{L}_{LSupCon}
\end{equation}

\subsection{Overview of the VABPP method}

After training the re-id model, we can use the VABPP method during testing process. The VABPP method can play the best performance in the steady state, and the more stable the state, the better the performance of the VABPP method. The steady state here refers to the statistical aspect, that is, the more data used for statistics, the better the performance of VABPP. We assume that the number of images per view is large enough for any vehicle and that most vehicles have images of all $V$ views, here $V$ is the total number of views in the dataset, so we make the following assumptions:

\begin{equation}
	\label{assumption}
	\frac{\sum_{p\in P_x^{v_s}}\frac{1}{\lvert P_x^{v_s} \rvert}d(f_x,f_p)}{k_{v,v_s}}\approx\frac{\sum_{p\in P_y^{v_t}}\frac{1}{\lvert P_y^{v_t} \rvert}d(f_y,f_p)}{k_{v,v_t}}
\end{equation}

\noindent Where $x$ and $y$ are arbitrary two images with the same view $v$ (also arbitrary), and $f_x, f_y$ and $f_p$ are the features of $x$, $y$ and $p$, respectively. $v_s$ and $v_t$ are any two views. $P_x^{v_s}$ and $P_y^{v_t}$ are the sets of vehicle images with the same vehicle ID as $x$ and $y$ respectively, and the views of the images in these two sets are $v_s$ and $v_t$ respectively. And  $\lvert\cdot\rvert$ represents the cardinality. $(i,j)$ represents a positive view pair when the view of the query image is i and the view of the positive image is j, $k_{i,j}\in K^{V\times V}$ is used to measure the difficulty of matching the positive image with the query image when the view of the query image is i and the view of the positive image is j. If the matching is relatively easy (i.e. the distance between the query image feature and the positive feature is small), the $k_{i,j}$ is also relatively small, on the contrary, the $k_{i,j}$ is relatively large. we stipulate $k_{i,i}=1, i=1,2,\dots,V$.

In formula \ref{assumption}, let $v_s=v_t$, then the following formula is obtained:

\begin{equation}
	\label{inference}
	\sum_{p\in P_x^{v_s}}\frac{1}{\lvert P_x^{v_s} \rvert}d(f_x,f_p)\approx\sum_{p\in P_y^{v_s}}\frac{1}{\lvert P_y^{v_s} \rvert}d(f_y,f_p)
\end{equation}
Since $x$ and $y$ have the same view, but not necessarily the same vehicle ID, so we have the following inference :

\textbf{Inference 1:} If vehicle image $x$ and $y$ have the same view, then the average of the distances between features of all positives of $x$ with view $v$ and the feature of $x$ and the average of the distances between features of all positives of $y$ with view $v$ and the feature of $y$ are almost equal.

We denote $c(\mathcal{V}(x),v_s)=\sum_{p\in P_x^{v_s}}\frac{1}{\lvert P_x^{v_s}\rvert}d(f_x,f_p)$ as the distance center of the view pair $(\mathcal{V}(x), v_s)$, where $\mathcal{V}(x)$ represents the function of finding the view ID of image $x$. $c(i,j)\in C^{V\times V}$, $C^{V\times V}$ is called view-pair distance center matrix.

Then let $v=v_t=i$ and $v_s=j$ in formula \ref{inference}, we get the following formula:

\begin{equation}
	\label{kij}
	\frac{c(i,i)}{c(i,j)}\approx\frac{1}{k(i,j)}
\end{equation}
Denote
\begin{equation}
	\label{delta}
	\delta(i,j)=\frac{1}{k(i,j)}\approx\frac{c(i,i)}{c(i,j)}
\end{equation}

We call $\delta(i,j)$ the view-pair distance scaling coefficient when the view of the query image is $i$ and the view of the candidate image is $j$. Here $\delta(i,j)\in \Delta^{V\times V}$, and $\Delta^{V\times V}$ is the view-pair distance scaling coefficient matrix. Similarly there are $\delta(i,i)=1, i=1,2,\dots, V$.
So far, according to formula \ref{delta}, we can get the following formula:
\begin{equation}
	\label{vabpp_core_formula}
	c(i,i)\approx c(i,1)\cdot \delta(i,1)\approx c(i,2)\cdot\delta(i,2)\approx\cdots\approx c(i,V)\cdot\delta(i,V)
\end{equation}
This formula shows that $\delta(i,j)$ can scale $c(i,j)$ to almost the same point as $c(i,i), (i,j=1,2,...V)$.

The inference process of formula \ref{vabpp_core_formula} is based on the test set, but the key problem here is that the test set cannot participate in the calculation of the matrix $\Delta^{V\times V}$, so we use the training set to calculate $\Delta^{V\times V}$, and finally apply it to the test set.

\begin{figure}[!t]
	\centering
	\includegraphics[width=.85\textwidth]{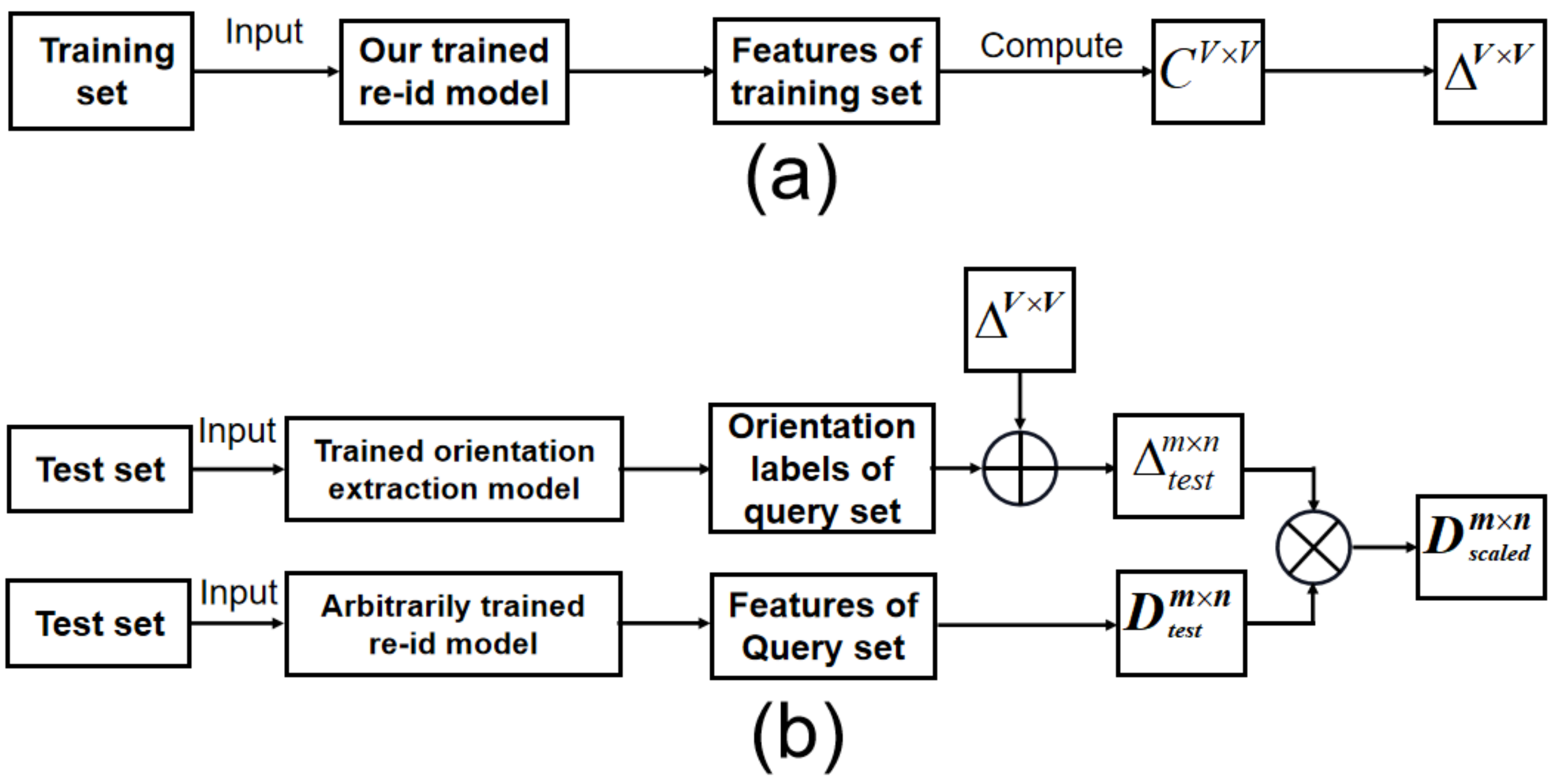}%
	\caption{The flow of the proposed VABPP method.}
	\label{vabpp_flow}
\end{figure}

The Fig \ref{vabpp_flow} shows the flow of the VABPP method. As can be seen in the Fig. \ref{vabpp_flow}(a), the VABPP method first uses the training set to calculate the view-pair distance center matrix $C^{V\times V}$, and then uses $C^{V\times V}$ to find $\Delta^{V\times V}$. The VABPP method can then be integrated into any other trained re-id model by directly using the matrix $\Delta^{V\times V}$ calculated in Fig. \ref{vabpp_flow}(a). As can be seen in the Fig. \ref{vabpp_flow}(b), if the VABPP method is directly applied to any other trained re-id model, we not only need the matrix $\Delta^{V\times V}$, but also a orientation extraction model to extract the orientation labels of all the test set images. So we provide the training set orientation labels of all the three datasets\footnote{Orientation labels can be found at: https://docs.zohopublic.com.cn/file/dfp\\vf58bf63e33c0a4d129eb2b753bd75b1cc}. We believe that if the orientation extraction model is not trained with the orientation labels we provided, the experimental results may be different and $\Delta^{V\times V}$ needs to be recalculated. Therefore, the rest of this section will first introduce the calculation process of $\Delta^{V\times V}$, and then introduce how to directly apply the VABPP method into other methods.

\subsection{The calculation of view-pair distance scaling coefficient matrix $\Delta^{V\times V}$}
\label{sec43}
Because the test set can not participate in the calculation of $\Delta^{V\times V}$, we use the training set to calculate $\Delta^{V\times V}$, and finally apply $\Delta^{V\times V}$ to the test set during the test process. Therefore, all the calculation of this section is based on the training set. As shown in Fig. \ref{network}, the specific method we calculate $\Delta^{V\times V}$ is as follows. Firstly we input all the images of the training set $T=\{t_1, t_2,\dots, t_l\}$ ($l$ is the total number of images in the training set) into our trained re-id network to extract features, then we obtain the feature set $F_T=\{f_1^T,f_2^T,\dots,f_l^T\}$ of the training set. However, the training set is not divided into query set $Q$ and gallery set $G$, so let $Q=T$ and $G=T$. Then their corresponding feature sets are $F_Q=F_T$, and $F_G=F_T$. For each query image $q\in Q$, we calculate the distances between the feature of $q$ and all its positive features in $F_G$, and use these distances to calculate the view-pair distance center matrix $C^{V\times V}$. Finally, we use $C^{V\times V}$ to calculate $\Delta^{V\times V}$.

\noindent Here, the calculation formula of element $c(i,j)$ of $C^{V\times V}$ is as follows:

\begin{equation}
	\label{cij}
	c_{i,j}=\frac{\sum_{q\in Q^(i)}\sum_{g\in G(j|q)}d(f_q,f_g)}{\sum_{q\in Q(i)}\sum_{g\in G(j|q)}1}
\end{equation}
\noindent Here, $Q(i)=\{q|q\in Q, \mathcal{V}(q)=i\}$ is the set of all images with view $i$ in $Q$, $G(j|q)=\{g|y(g)=y(q),\mathcal{V}(g)=j,Camid(g)\neq Camid(q)\}$ is a set of all images in $G$ that have the same vehicle ID as the query image $q$ and have view $j$, but all images with the same camera as $q$ are removed.$f_q$ and $f_g$ represent the features of image $q$ and $g$, respectively. According to formula \ref{delta}, the calculation formula of element $\delta(i,j) (i,j=1,2,\dots, V)$ of  is as follows:

\begin{equation}
	\label{delta2}
	\delta_{i,j}=\left\{
	\begin{array} {ll}
		1,&i=j \\
		c_{i,j}/c_{i,i},&i\ne j
	\end{array} \right.
\end{equation}

We can know from formula \ref{delta2} that $\Delta^{V\times V}$ is a matrix whose main diagonal elements are all 1, indicating that before and after the application of VABPP method the distances between the features of the images with the same view as the query image (i.e. $i=j$) and the feature of the query image does not change.

\begin{figure}[!t]
	\centering
	\includegraphics[width=.85\textwidth]{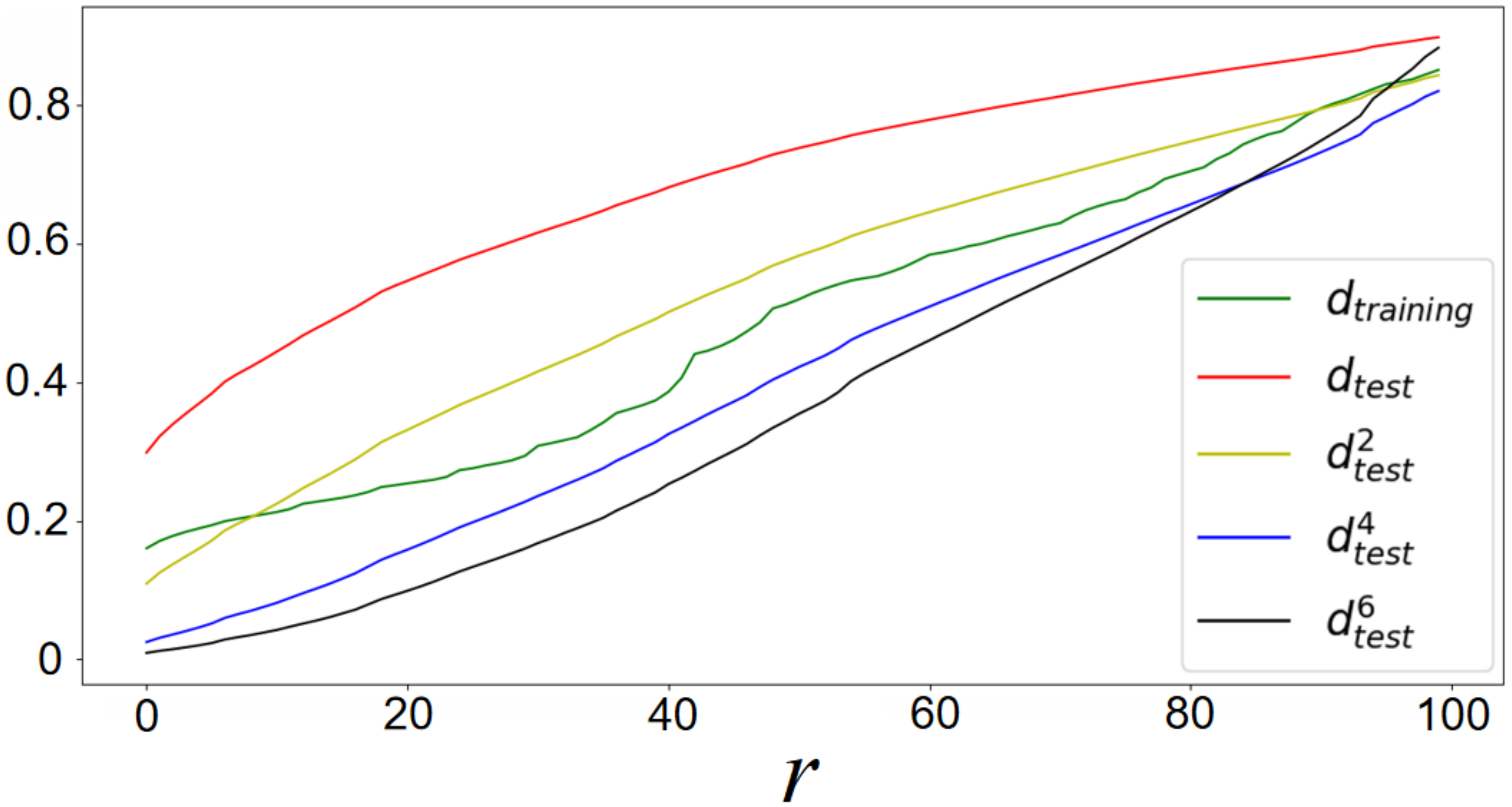}%
	\caption{Different distance distributions between training set and test set. All curves are drawn according to the formula $d(r)=\frac{1}{\lvert Q\rvert}\sum_{q\in Q}d^\alpha_{sorted}(r|q)$. Where $\alpha$ is a hyper exponential parameter and $Q$ is the query set, $d_{sorted}(r|q)$ is the $r$-th value of the distance between the query image $q$ and all the images in gallery set $G$ in ascending order. For the drawing of curve $d_{training}$, we set $Q=G=T$, where $T$ is the training set. All these curves are drawn under the condition of removing the images with the same ID and same camera as the query image $q$ in the gallery set. It is obvious that  $d_{training}$,  $d_{test}^4$ and $d_{test}^6$ are concave curves, while $d_{test}$ and $d_{test}^2$ are convex curves.}
	\label{distance_distribution2}
\end{figure}

\subsection{Modify test distance}

After calculating $\Delta^{V\times V}$ from the training set, we are not in a hurry to use it to modify the test distance (MTD), because the test set and training set have different distance distributions and we must unify the distance distributions between the test set and the training set. Fig. \ref{distance_distribution2} shows the difference between the distance distribution of the training set and the test set of our baseline. As can be seen from Fig. \ref{distance_distribution2}, the distance distribution of the test set (i.e. $d_{test}$) is a smooth convex curve, and the distance distribution of the training set (i.e. $d_{training}$) is a curve from concave to convex. The reason is that baseline uses $\mathcal{L}_{LSupCon}$ to pull the positive features closely and push the negative features away during training, but there is no such mechanism during testing. Since it can be observe from formula \ref{assumption} we are more concerned about the positive features, in order to make the distances between the positive features and query features in the test set have similar distance distribution with the distances between the positive features and anchor features in the training set, in the test set, for any $q\in Q$ and $g\in G$, we make the following distance modifications:
\begin{equation}
	\label{d_unified}
	d_{unified}(f_q,f_g|\theta)=d^{\gamma}(f_q,f_g|\theta)
\end{equation}

\noindent Where $\gamma$ is a hyper exponential parameter. $d(f_q,f_g|\theta)\in D^{m\times n}_{test}$, $D^{m\times n}_{test}$ is the matrix composed of the distance between the features of all images in the query set $Q$ and the features of all the images in the gallery set $G$, $d_{unified}(f_q,f_g|\theta)\in D^{m\times n}_{unified}$, $D^{m\times n}_{unified}$ is the distance matrix composed of all elements of  $D^{m\times n}_{test}$ applied by the UDD method. 

We do this for the following two reasons. First, by observing the distances distributions in the baseline (Fig. \ref{distance_distribution2}) , we find that the distances between positive pairs is far less than 1 in both the training set and the test set. Second, the distance between the positive pairs in the test set is generally larger than that in the training set. The value of $\gamma$ can change the bending degree and bending direction of the distance curves in the test set.

\textbf{Note:} in Fig. \ref{distance_distribution2}, it seems that the curve of $d^6_{test}$ is the most similar to that of $d_{training}$, but the fact is not necessarily when $\gamma=6$ is the best. The reasons are as follows:

(1) Several loss functions are used in the training of the training set, and the training accuracy is very high. Almost all the top ranking images are positive images.

(2) When drawing these curves, the distances between the feature of each query image and the features of all images in the gallery are sorted from small to large, then all such distances calculated by all query images are averaged at the corresponding position. However, the test accuracy is not as high as the training accuracy, and the test distances are far larger than the training distances. Therefore, the $n$-th power in Fig. \ref{distance_distribution2} will be disturbed by the larger value.

(3) The number of positive images in the test set is different from that in the training set, so that the abscissa in Fig. \ref{distance_distribution2} cannot be aligned in the test set and training set.

Now, we start to modify the test distance. In Section \ref{sec32}, we analyze that for a query image $q$ with view $i$, $\delta(i,j)$ can scale $c(i,j)$ to the same point as $c(i,i) (i,j=1,2,\dots,V)$. But what we need is an overall scaling of the distances between all gallery image features with view $j$ and the query image, so we need to modify all test distances. The method of modifying test distance is abbreviated as MTD. The goal of the MTD method is to use the following distance formula to conduct the final re-id similarity matching:

\begin{equation}
	\label{d_scaled}
	d_{scaled}(f_q,f_g|\theta)=d_{unified}(f_q,f_g|\theta)\times\delta(\mathcal{V}(q),\mathcal{V}(g))
\end{equation}

\noindent Where, $q\in Q$ and $g\in G$, $Q=\{q_1,q_2,\dots,q_m\}$ is the query set, $G=\{g_1,g_2,\dots,g_n\}$ is the gallery set, and $\theta$ is the model parameter. $d_{scaled}(f_q,f_g|\theta)\in D^{m\times n}_{scaled}$. $D^{m\times n}_{scaled}$ is a distance matrix composed of all elements of $D^{m\times n}_{unified}$ applied by the UDD method.

\subsection{Steps for integrating VABPP into other trained re-id model}

Using the  calculated in this paper and the orientation labels of the training set given in this paper, the steps of integrating VABPP method into other methods are as follows:

Step1. Train a re-id model and a orientation extraction network with specific methods

Step2. Extract the vehicle features and view labels of the test set.

Step3. Normalize vehicle features and calculate the distances between all query image features and image features of all gallery sets to form a distance matrix $D^{m\times n}_{test}$, where m and n represent the number of images in the query set and gallery set respectively.

Step4. Use formula 16 to unify the test distance distribution and training distance distribution to obtain the unified distance matrix $D^{m\times n}_{unified}$ of the test set.

Step5. Use the extracted test set view labels and $\Delta^{V\times V}$ to calculate the view-pair distance scaling coefficient matrix $\Delta^{m\times n}_{test}$ between all images of the query set and images of gallery sets.

Step6. Use $D^{m\times n}_{unified}\bigcdot \Delta^{m\times n}_{test}$ to replace the original $D^{m\times n}_{test}$ for similarity matching.

\section{Experiments}
\subsection{Datasets}
\textbf{VeRi-776 dataset} There are 20 cameras were used to take images of this dataset, and images of each vehicle are taken by 2-18 cameras. The training set contains 37778 images of 576 vehicles. The query set and gallery set contain the same 200 vehicle IDs, and the number of images are 1678 and 11579 respectively. There are eight orientations in this dataset, including front, rear, left, right, left front, left rear, right front and right rear.

\textbf{VehicleID dataset} This dataset has no camera information, and all images are taken from the front or rear. The training set consists of 113346 images of 13164 vehicles. The test set is divided into three sub sets: large, medium and small, including 19777, 13377 and 6493 images of 2400, 1600 and 800 vehicles respectively. Randomly select one image from each vehicle in these subsets to form the corresponding gallery set, and the rest form the corresponding query set.

\textbf{VERI\_Wild dataset} There are 174 cameras were used to take images of this dataset. The training set contains 277797 images of 30671 vehicles. The test set is divided into three subsets: large, medium and small. The query set of the three subsets contains 3000, 5000 and 10000 vehicle images (one image per vehicle), and the gallery set contains 38861, 64389 and 128517 images respectively. This dataset includes six orientations: front, rear, left front, left rear, right front and right rear.

\subsection{Training configurations}
We use Huynh \cite{ref24} as our baseline and resnext101\_ibn\_a as the backbone. But we removed the mixstyle module of the baseline. We use the Momentum Adaptive Loss Weight method introduced in references \cite{ref24,ref40} to update the weights $\lambda_{ID}$ and $\lambda_{Metrix}$. We resize the image size to 320$\times$ 320, and apply data enhancement methods such as color jitters, random flip, brightness and contrast adjustment, random erase and random cropping. The batch size is 64. We use ADAM optimizer with the cosine annealing scheduler, total training epoch is set to 24. For the VeRi-776 dataset, the batch is composed of 8 identities, each identity contains 8 images, and the initial learning rate is set to 3.5$\times$ 10-4. For the VehicleID dataset, the batch is composed of 16 identities, each identity contains 4 images, and the initial learning rate is set to 3.5$\times$ 10-5. For VERI\_Wild dataset, the batch is composed of 32 identities, each identity contains 2 images, and the initial learning rate is set to 10-4. The program is implemented on pytorch, and use a single NVIDIA GeForce RTX 3090 GPU.

\begin{table*}[htb]
\begin{tiny}
	\caption{In the VeRi-776 dataset, VehicleID dataset and VERI\_Wild dataset, we use $\mathcal{L}_{LSupCon} (\mathcal{L}_L)$ and $\mathcal{L}_{GSupCon} (\mathcal{L}_G)$ and the combination of these two losses, and respectively use resnet50\_ibn\_a,resnext101\_ibn\_a and resnet152 as the backbones. r50, r101 and r152 respectively represent resnet50\_ibn\_a,resnext101\_ibn\_a and resnet152.}\label{tab1}
	\centering
	\setlength{\tabcolsep}{0.06mm}
	\begin{tabular}{|c|c|c|c|c|c|c|c|c|c|c|c|c|c|c|c|}
		\hline
		\multirow{3}{*}{Backbone}&\multirow{3}{*}{Method}&\multicolumn{2}{c|}{\multirow{2}{*}{VeRi-776}}& \multicolumn{6}{c|}{VehicleID}& \multicolumn{6}{c|}{VERI\_Wild}\\
		\cline{5-16}
		\multicolumn{1}{|c|}{}&\multicolumn{1}{c|}{}&\multicolumn{2}{c|}{}&\multicolumn{2}{c|}{small}&\multicolumn{2}{c|}{medium}&\multicolumn{2}{c|}{large}&\multicolumn{2}{c|}{small}&\multicolumn{2}{c}{medium}&\multicolumn{2}{|c|}{large} \\
		\cline{3-16}
		\multicolumn{1}{|c|}{}&\multicolumn{1}{c|}{}&mAP(\%)&r1(\%)&r1(\%)&r5(\%)&r1(\%)&r5(\%)&r1(\%)&r5(\%)&mAP(\%)&r1(\%)&mAP(\%)&r1(\%)&mAP(\%)&r1(\%)\\
		\hline
		\multirow{3}{*}{r50}&$\mathcal{L}_L$&79.6&96.5&80.0&96.3&76.2&93.5&73.2&90.8&76.6&91.1&70.1&87.6&61.3&82.8 \\
		\cline{2-16}
		\multicolumn{1}{|c|}{}&$\mathcal{L}_G$&79.3&96.5&80.4&96.2&77.0&92.8&74.6&90.2&\textbf{84.8}&\textbf{94.0}&\textbf{79.1}&\textbf{91.5}&\textbf{71.1}&\textbf{87.6}\\
		\cline{2-16}
		\multicolumn{1}{|c|}{}&$\mathcal{L}_L + \mathcal{L}_G$&\textbf{80.8}&\textbf{96.6}&\textbf{82.8}&\textbf{97.9}&\textbf{79.5}&\textbf{95.4}&\textbf{76.2}&\textbf{92.4}&84.1&92.9&78.7&90.1&71.0&85.4\\
		\hline
		\multirow{3}{*}{r101}&$\mathcal{L}_L$&81.2&96.5&80.9&97.0&76.5&92.6&73.9&91.0&78.7&91.9&72.8&89.3&64.3&84.8 \\
		\cline{2-16}
		\multicolumn{1}{|c|}{}&$\mathcal{L}_G$&81.0&96.7&82.0&96.9&78.1&94.0&75.7&91.2&\textbf{86.6}&\textbf{94.8}&\textbf{81.9}&\textbf{93.3}&\textbf{74.7}&\textbf{89.6}\\
		\cline{2-16}
		\multicolumn{1}{|c|}{}&$\mathcal{L}_L + \mathcal{L}_G$&\textbf{83.2}&\textbf{97.3}&\textbf{85.8}&\textbf{98.0}&\textbf{80.1}&\textbf{96.0}&\textbf{78.3}&\textbf{93.7}&84.7&93.3&79.0&89.8&70.7&84.9\\
		\hline
		\multirow{3}{*}{r152}&$\mathcal{L}_L$&79.0&96.5&80.2&96.7&76.0&94.2&73.0&91.6&76.3&90.4&70.3&87.7&61.2&82.2 \\
		\cline{2-16}
		\multicolumn{1}{|c|}{}&$\mathcal{L}_G$&79.1&96.5&83.0&96.9&78.6&94.4&76.0&91.7&\textbf{83.7}&\textbf{93.3}&\textbf{78.1}&\textbf{91.4}&\textbf{70.1}&\textbf{86.8}\\
		\cline{2-16}
		\multicolumn{1}{|c|}{}&$\mathcal{L}_L + \mathcal{L}_G$&\textbf{80.8}&\textbf{96.6}&\textbf{84.4}&\textbf{97.9}&\textbf{81.0}&\textbf{96.2}&\textbf{76.9}&\textbf{93.5}&82.6&91.2&76.6&87.6&67.6&81.8\\
		
		\hline
	\end{tabular}
\end{tiny}
\end{table*}

\subsection{Relationship between $\mathcal{L}_{LSupCon}$ and $\mathcal{L}_{GSupCon}$}

As mentioned in section \ref{sec_lgsupcon}, $\mathcal{L}_{GSupCon}$ is better for larger datasets. For smaller datasets, we need to combine $\mathcal{L}_{LSupCon}$ and $\mathcal{L}_{GSupCon}$ to train the model, we use three datasets of different sizes to verify this conclusion. The VeRi-776 dataset contains 576 training IDs, the VehicleID dataset contains 13164 training IDs, and the VERI\_Wild dataset contains 30671 training IDs. Further more, in order to show the generalization of this conclusion, we carry out corresponding experiments on three backbone models: resnet50\_ibn\_a,resnext101\_ibn\_a and resnet152. TABLE \ref{tab1} shows the experimental results. 

\textbf{VeRi-776 dataset} In TABLE \ref{tab1}, we find that in the VeRi-776 dataset, the experimental results of $\mathcal{L}_{GSupCon}$ are not significantly better than that of $\mathcal{L}_{LSupCon}$, but the results of $\mathcal{L}_{LSupCon}$ + $\mathcal{L}_{GSupCon}$ are significantly better than that of any one of the two losses used alone, indicating that the number of training IDs (or images) for this dataset is not sufficient for the proposed $\mathcal{L}_{GSupCon}$ to perform well.

\textbf{VehicleID dataset} The number of training IDs in the VehicleID dataset is significantly larger than that in the VeRi-776 dataset. And we can find in TABLE \ref{tab1}, in this dataset, the results of using $\mathcal{L}_{GSupCon}$ for the three backbones are better than those using $\mathcal{L}_{LSupCon}$. In the large sub test set, the rank-1 has been improved by 3\% for the resnet152 backbone. And the experimental results of $\mathcal{L}_{LSupCon}$ + $\mathcal{L}_{GSupCon}$ are also significantly better than that of using either of the two losses alone. Compared with baseline (i.e. $\mathcal{L}_{LSupCon}$), the improvement is more significant, and the rank-1 has been improved by 4.9\% for the resnet101\_ibn\_a backbone in the small sub test set, 
indicating that the number of training IDs (or images) for this dataset is large enough to make the performance of the proposed $\mathcal{L}_{GSupCon}$ used alone to exceed that of the $\mathcal{L}_{LSupCon}$ used alone.

\textbf{VERI\_Wild dataset} For this dataset, as the number of training IDs in this dataset reaches 30671, which is the largest in the three dataset, it can be seen that with the increase of the number of positive images and negative IDs, the mAPs of using $\mathcal{L}_{GSupCon}$ in the three backbones than those of using $\mathcal{L}_{LSupCon}$ (i.e. baseline) with an increase of almost all by 8\%. Even in the large sub test set, for resnet101\_ibn\_a backbone, the mAP of using $\mathcal{L}_{GSupCon}$ has an increase of more than 10\% compared with that of using $\mathcal{L}_{LSupCon}$. For this data set, the results of using $\mathcal{L}_{GSupCon}$ is obviously better than those of using $\mathcal{L}_{LSupCon}$ + $\mathcal{L}_{GSupCon}$, which shows that the number of training IDs of this dataset is large enough, the $\mathcal{L}_{GSupCon}$'s one-way movement mode which tends to move to the global optimal solution has played a good role, and this shows that when $\mathcal{L}_{LSupCon}$ + $\mathcal{L}_{GSupCon}$ is used to train the model, the local-to-local two-way movement mode of $\mathcal{L}_{LSupCon}$ affects the performance of $\mathcal{L}_{GSupCon}$.

\subsection{The ablation experiment of view-aware based post processing method}

We use resnext101\_ibn\_a as the backbone to conduct ablation experiments on MTD method and on the value of $\gamma$ for UDD method. For the VeRi-776 dataset and VehicleID dataset, we use $\mathcal{L}_{LSupCon}$ + $\mathcal{L}_{GSupCon}$ as metric loss, while for the VERI\_Wild dataset, we only use $\mathcal{L}_{GSupCon}$ as the metric loss. As can be seen in TABLE \ref{tab2}, when $\gamma=2$, the result is the best, so the value of $\gamma$ in subsequent experiments is 2. Since all the curves in Fig. \ref{distance_distribution2} are the results of sorting the distances between the feature of each query image and the features of all images in the gallery from small to large, then all such distances calculated by all query images are averaged at the corresponding positions, while the training set is well trained, but the test set are not trained, although the bending direction and bending degree of the distance distribution curves of $d^6_{test}$ (i.e. $\gamma=6$) and the distance distribution curve of training set (i.e. $d_{training}$) are the closest in Fig. \ref{distance_distribution2}, however, this does not mean that the experimental result is the best when $\gamma=6$. Fig. \ref{distance_distribution2} can only show that the distance distributions between the test set and the training set are different.

\begin{table*}[htb]
	\begin{tiny}

	\caption{The ablation experiments of VABPP method in the three datasets.}\label{tab2}
	\centering
	\setlength{\tabcolsep}{0.3mm}
	\begin{tabular}{|c|c|c|c|c|c|c|c|c|c|c|c|c|c|c|}
		\hline
		\multirow{3}{*}{Method}&\multicolumn{2}{c|}{\multirow{2}{*}{VeRi-776}}& \multicolumn{6}{c|}{VehicleID}& \multicolumn{6}{c|}{VERI\_Wild}\\
		\cline{4-15}
		\multicolumn{1}{|c|}{}&\multicolumn{2}{c|}{}&\multicolumn{2}{c|}{small}&\multicolumn{2}{c|}{medium}&\multicolumn{2}{c|}{large}&\multicolumn{2}{c|}{small}&\multicolumn{2}{c}{medium}&\multicolumn{2}{|c|}{large} \\
		\cline{2-15}
		\multicolumn{1}{|c|}{}&mAP(\%)&r1(\%)&r1(\%)&r5(\%)&r1(\%)&r5(\%)&r1(\%)&r5(\%)&mAP(\%)&r1(\%)&mAP(\%)&r1(\%)&mAP(\%)&r1(\%)\\
		\hline
		w/o MTD&83.2&97.3&85.8&98.0&80.1&96.0&78.3&93.7&86.6&94.8&81.9&93.3&74.7&89.6\\
		\hline
		UDD ($\gamma=1$)&83.4&96.9&85.9&\textbf{99.0}&81.1&97.1&76.8&\textbf{95.3}&87.3&94.2&82.7&91.8&75.5&87.7\\
		\hline
		UDD ($\gamma=2$)&\textbf{83.9}&\textbf{97.1}&\textbf{87.3}&98.8&\textbf{82.8}&\textbf{97.2}&\textbf{79.9}&95.0&\textbf{87.7}&94.8&\textbf{83.2}&92.8&\textbf{76.3}&89.3\\
		\hline
		UDD ($\gamma=3$)&83.7&\textbf{97.1}&87.1&98.6&82.3&97.0&79.7&94.8&87.5&94.8&83.0&93.1&76.1&89.5\\
		\hline
		UDD ($\gamma=4$)&83.6&\textbf{97.1}&86.7&98.6&81.9&96.8&79.5&94.6&87.3&\textbf{94.9}&82.8&93.2&75.8&89.5\\
		\hline
		UDD ($\gamma=5$)&83.6&\textbf{97.1}&86.6&98.4&81.6&96.5&79.3&94.4&87.2&94.8&82.6&93.2&75.6&89.5\\
		\hline
		UDD ($\gamma=6$)&83.5&\textbf{97.1}&86.4&98.4&81.4&96.5&79.2&94.3&87.2&\textbf{94.9}&82.5&\textbf{93.3}&75.5&\textbf{89.6}\\
		
		\hline
	\end{tabular}
\end{tiny}
\end{table*}

\begin{figure*}[!t]
	\centering
	\includegraphics[width=\textwidth]{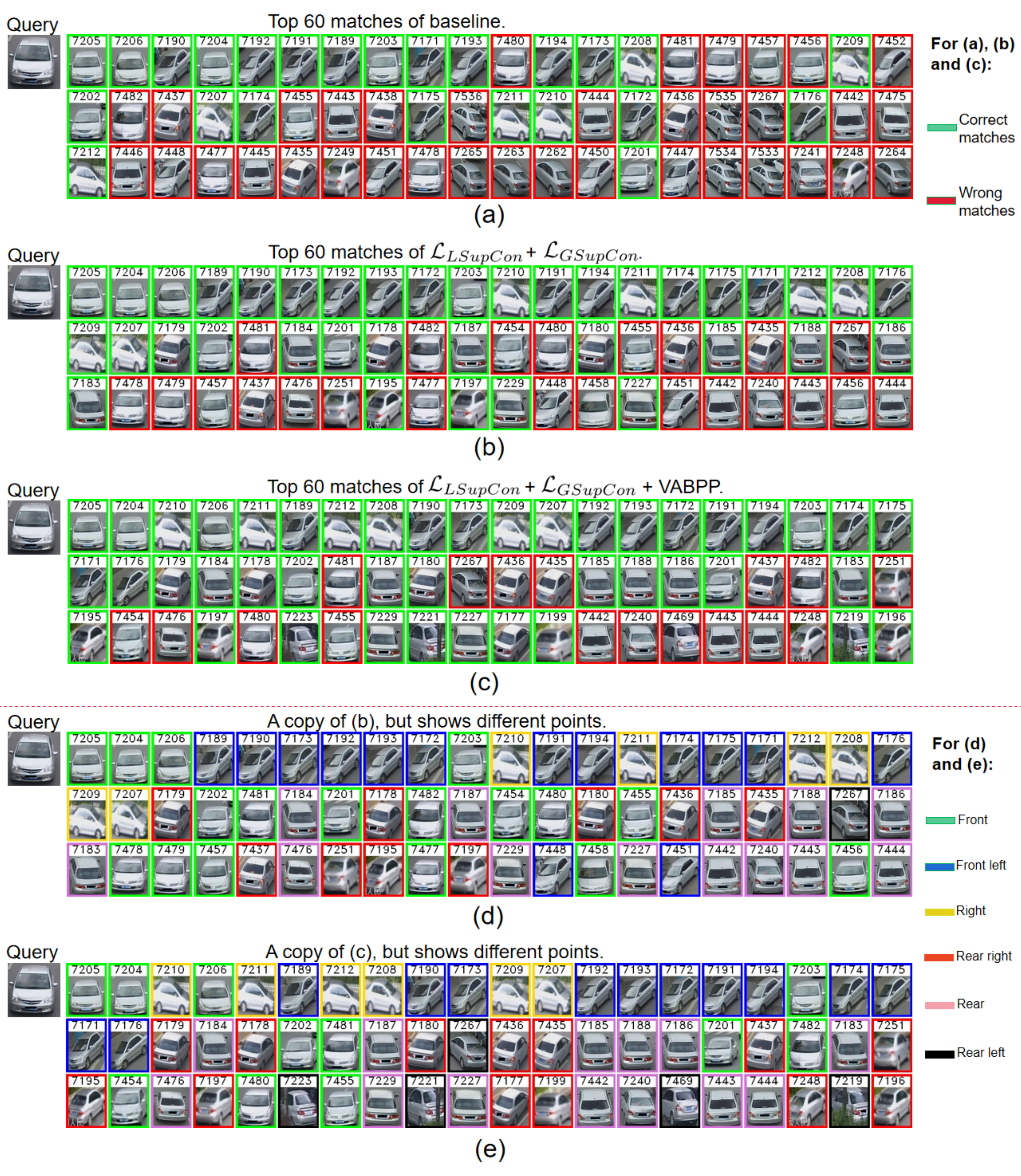}%
	\caption{Demonstration of the effect of the proposed method. The query images in (a), (b), (c), (d) and (e) are the same image. The number in each box represents the original index of the image in this box in the gallery set.}
	\label{demo_final}
\end{figure*}

\begin{table}[htb]
	\caption{In Bag-of-Tricks \cite{ref41}, the comparison of before and after applying VABPP method. The $\Delta ^{V\times V}$ in TABLEs \ref{tab5}, \ref{tab6} and \ref{tab7} in the Appendix B are directly used here. "†" in the table indicates the VABPP method is used, where $\gamma=4$.}\label{tab3}
	\centering
	\setlength{\tabcolsep}{3.85mm}
	\begin{tabular}{|c|c|c|c|c|}
		\hline
		\multicolumn{2}{|c|}{Dataset}&mAP&rank-1&rank-5\\
		\hline
		\multicolumn{2}{|c|}{VeRi-776}&77.1&95.4&98.2\\
		\hline
		\multicolumn{2}{|c|}{VeRi-776†}&\textbf{78.6}&\textbf{95.3}&\textbf{98.2}\\
		\hline
		\multirow{6}{*}{VehicleID}&Small&89.0&83.6&96.0\\
		\cline{2-5}
		\multicolumn{1}{|c|}{}&Small†&\textbf{92.3}&\textbf{87.8}&\textbf{97.9}\\
		\cline{2-5}
		\multicolumn{1}{|c|}{}&Medium&84.8&78.7&93.2\\
		\cline{2-5}
		\multicolumn{1}{|c|}{}&Medium†&\textbf{89.8}&\textbf{85.0}&\textbf{96.0}\\
		\cline{2-5}
		\multicolumn{1}{|c|}{}&Large&83.6&77.7&91.6\\
		\cline{2-5}
		\multicolumn{1}{|c|}{}&Large†&\textbf{87.8}&\textbf{82.6}&\textbf{94.4}\\
		\hline
		\multirow{6}{*}{VERI\_Wild}&Small&77.0&92.1&97.5\\
		\cline{2-5}
		\multicolumn{1}{|c|}{}&Small†&\textbf{80.5}&\textbf{92.3}&\textbf{97.6}\\
		\cline{2-5}
		\multicolumn{1}{|c|}{}&Medium&70.9&89.4&95.8\\
		\cline{2-5}
		\multicolumn{1}{|c|}{}&Medium†&\textbf{74.7}&\textbf{89.6}&\textbf{96.0}\\
		\cline{2-5}
		\multicolumn{1}{|c|}{}&Large&62.6&85.3&93.4\\
		\cline{2-5}
		\multicolumn{1}{|c|}{}&Large†&\textbf{66.7}&\textbf{85.5}&\textbf{93.6}\\

		\hline
	\end{tabular}
\end{table}

\begin{table*}[htb]
	\begin{tiny}
		
	\caption{Comparison between the proposed method and the state-of-the-art methods in the three datasets. m(\%) meas mAP(\%).}\label{tab4}
	\centering
	\setlength{\tabcolsep}{0.000001mm}
	\begin{tabular}{|c|c|c|c|c|c|c|c|c|c|c|c|c|c|c|c|}
		\hline
		\multirow{3}{*}{method}&\multicolumn{2}{c|}{\multirow{2}{*}{VeRi-776}}& \multicolumn{6}{c|}{VehicleID}& \multicolumn{6}{c|}{VERI\_Wild}\\
		\cline{4-15}
		\multicolumn{1}{|c|}{}&\multicolumn{2}{c|}{}&\multicolumn{2}{c|}{small}&\multicolumn{2}{c|}{medium}&\multicolumn{2}{c|}{large}&\multicolumn{2}{c|}{small}&\multicolumn{2}{c}{medium}&\multicolumn{2}{|c|}{large} \\
		\cline{2-15}
		\multicolumn{1}{|c|}{}&m(\%)&r1(\%)&r1(\%)&r5(\%)&r1(\%)&r5(\%)&r1(\%)&r5(\%)&m(\%)&r1(\%)&m(\%)&r1(\%)&m(\%)&r1(\%)\\
		\hline
		VVAER \cite{ref13}&61.2&89.0&74.7&93.8&68.6&90.0&63.5&85.6&62.2&75.8&53.7&68.2&41.7&58.7\\
		\hline
		PCRNet \cite{ref42}&78.6&95.4&86.6&98.1&82.2&96.3&80.4&94.2&81.2&92.5&75.3&89.6&67.1&85.0\\
		\hline
		VARID \cite{ref36}&79.3&96.0&85.8&96.9&81.2&94.1&79.5&92.2&75.4&75.3&70.8&68.8&64.2&63.2\\
		\hline
		VSCR \cite{ref43}&75.5&94.1&74.6&87.1&-&-&-&-&75.8&93.1&70.5&89.7&64.2&86.3\\
		\hline
		VAT \cite{ref44}&80.4&97.5&84.5&-&80.5&-&78.2&-&-&-&-&-&-&-\\
		\hline
		\hline
		
		GSTE \cite{ref20}&59.5&96.2&75.9&84.2&74.8&83.6&74.0&82.7&31.4&60.5&26.2&52.1&19.5&45.4\\
		\hline
		FDA-Net\cite{ref22}&55.5&84.3&-&-&59.9&77.1&55.5&74.7&35.1&64.0&29.8&57.8&22.8&49.4\\
		\hline
		SAVER \cite{ref45}&79.6&96.4&79.9&95.2&77.6&91.1&75.3&88.3&80.9&94.5&75.3&92.7&67.7&\textbf{89.5}\\
		\hline
		HRCN\cite{ref9}&83.1&\textbf{97.3}&\textbf{88.2}&98.4&81.4&96.6&\textbf{80.2}&94.4&85.2&94.0&80.0&91.6&72.2&88.0\\
		\hline
		LCDNet+BRL+RR \cite{ref46}&82.3&96.1&85.6&97.2&79.0&94.2&75.2&90.6&-&-&-&-&-&-\\
		\hline
		\hline
		Baseline \cite{ref24}&81.2&96.5&80.9&97.0&76.5&92.6&73.9&91.0&78.7&91.9&72.8&89.3&64.3&84.8\\
		\hline
		$\mathcal{L}_L+\mathcal{L}_G(ours)$&\textbf{83.9}&97.1&87.3&\textbf{98.8}&\textbf{82.8}&\textbf{97.2}&79.9&\textbf{95.0}&-&-&-&-&-&-\\
		\hline
		$\mathcal{L}_G(ours)$&-&-&-&-&-&-&-&-&\textbf{87.7}&\textbf{94.8}&\textbf{83.2}&\textbf{92.8}&\textbf{76.3}&\textbf{89.3}\\

		\hline
	\end{tabular}

\end{tiny}
\end{table*}
\subsection{Result demonstration}

We present a demonstration result of the proposed method, which is based on the model trained by $\mathcal{L}_{GSupCon}+\mathcal{L}_{LSupCon}$. And because in VeRi-776 dataset, each vehicle contains 65.59 images in average, it is very suitable for demonstrating, so we choose a image from this dataset to do this experiment. The filename of the query image used to demonstrate is "0482\_c011\_00052840\_0.jpg". Comparing Fig. \ref{demo_final}(a) and Fig. \ref{demo_final}(b), it can be seen that the matching result of the $\mathcal{L}_{LSupCon}+\mathcal{L}_{GSupCon}$ is much better than that of the baseline $(\mathcal{L}_{LSupCon})$. The top 24 images in Fig. \ref{demo_final}(b) are all correct matches, but there are 8 wrong matches in the top 24 images in Fig. \ref{demo_final}(a), indicating that increasing the global attribute plays a very important and positive role in improving the accuracy.

Comparing Fig. \ref{demo_final}(b) and Fig. \ref{demo_final}(c), we can see that some rankings of the negative images with the same view as the query image (front view), such as images with indexes 7481, 7482, 7454, 7480, have been moved back a lot in Fig. \ref{demo_final}(c) compared with Fig. \ref{demo_final}(b), because their distance scaling coefficient is the largest (first row of TABLE \ref{tab5}, $\delta(0,0)=1$), and the distances between the features of images of other views and the feature of the query image become smaller, but the distances between the features of the images of the front view and the feature of the query image are invariable, so that the rankings of almost all images with front view become more rearward. Meanwhile, there are only 37 correctly matches in Fig. \ref{demo_final}(b), but there are a total of 43 correctly matches in Fig. \ref{demo_final}(c). It can be seen that the VABPP method is very helpful to improve the experimental accuracy. The main reason why VABPP method is effective is that it can make the rankings of positive images that are difficult to match higher.

Fig. \ref{demo_final}(d) and Fig. \ref{demo_final}(e) show why the VABPP method is effective. It can be seen that although the overall rankings of Fig. \ref{demo_final}(d) and Fig. \ref{demo_final}(e) is different, the rankings within images of the same color (i.e. the same view) is fixed, and the VABPP method increases the cross-ranking between positives with different views.

\subsection{Application examples of VABPP method}
We list all the $\Delta^{V\times V}$s matrices calculated by our trained re-id model on the VeRi-776 dataset, VehicleID dataset, and VERI\_Wild dataset in TABLE \ref{tab5}, TABLE \ref{tab6} and TABLE \ref{tab7} in Appendix B, respectively. Recently, the most commonly used baseline in the vehicle re-id field is Bag-of-Tricks \cite{ref41}. In order to verify that the VABPP method can be easily integrated into other methods, in the three datasets, we apply the VABPP method to the Bag-of-Tricks \cite{ref41}. For this experiment, we don’t need to recalculate $\Delta^{V\times V}$. For these three datasets, we directly use the corresponding $\Delta^{V\times V}$ in the TABLE \ref{tab5}, TABLE \ref{tab6} and TABLE \ref{tab7}. The experimental results of Bag-of-Tricks \cite{ref41} and Bag-of-Tricks \cite{ref41}+VABPP are shown in TABLE \ref{tab3}. As can be seen from TABLE \ref{tab3}, all the results of Bag-of-Tricks \cite{ref41} + VABPP are much higher than that of without VABPP method in all the three datasets. In the VeRi-776 dataset, the mAP of with VABPP is 1.48\% higher than that of without VABPP. In the three sub test sets of the VehicleID dataset, CMC@1s of with VABPP are 3.23\%, 5.03\% and 4.15\% higher than those of without VABPP, respectively. In the three sub test sets of the VERI\_Wild dataset, mAPs of with VABPP are 3.48\%, 3.83\% and 4.10\% higher than that of without VABPP, respectively. It can be seen that such a great improvement is achieved at the condition of without affecting the training process and without recalculating matrix $\Delta^{V\times V}$, indicating that the proposed method has a high practical value. The Bag-of-Tricks \cite{ref41} is a commonly used baseline before, so we believe that the VABPP method can be easily integrated into other methods by directly using the matrices $\Delta^{V\times V}$s we calculated. It should be emphasized that in TABLE \ref{tab2}, $\gamma=2$ is the best, while in TABLE \ref{tab3}, it is the best when $\gamma=4$. The reason is that the relationship between the distance distributions of training set and test set are different when the model structures are different or training methods are different, so it is necessarily to re-verify the optimal value of $\gamma$. In addition, it is important to ensure that test features are normalized.

By analyzing the results in TABLE \ref{tab2} and TABLE \ref{tab3}, we can also see that the performances of VABPP in VehicleID dataset and VERI\_Wild dataset are significantly better than that in VeRi-776 dataset. The reasons are as follows:

1)The total number of training images in the VeRi-776 dataset is 37778, and there are 64 values need to be counted (TABLE \ref{tab5}). While the total number of training images in the VehicleID dataset and VERI\_Wild dataset are 113346 and 277797 respectively, there are only 4 (TABLE \ref{tab6}) and 36 (TABLE \ref{tab7}) values need to be counted respectively, which makes the statistical values in the later two datasets are more general.

2)Although we divid the VeRi-776 dataset into 8 views, in fact, this dataset is far more than 8 views, because this dataset can be divided into 8 views in the horizontal position and can also be continuously divided in the vertical position. The reason is that some of the cameras in this dataset are in high positions and some are in low positions, which makes the division of only 8 views is significantly too few. The reason why we no longer continue to make more detailed division is that the total number of images in this dataset is too small, and the calculation of the $\Delta^{V\times V}$ matrix needs to be done in view-pairs of images within the same vehicle ID. If we continue to make detailed division, due to the small size of VeRi-776 dataset, the  $\Delta^{V\times V}$ will be even less statistically significant. While the VehicleID dataset and VERI\_Wild dataset have only two and six orientations respectively, and all the cameras used to capture these two datasets are almost at the same horizontal position.

\subsection{Compare with the state-of-the-art methods}
We compare the proposed methods with some state-of-the-art methods in the VeRi-776 dataset, VehicleID dataset and VERI\_Wild dataset. These methods include:

(1) View-aware based method: include VVAER \cite{ref13}, PCRNet \cite{ref42}, VARID \cite{ref36}, VSCR \cite{ref43} and VAT \cite{ref44}

(2) Metric learning based method: include GSTE \cite{ref20}, FDA-Net \cite{ref22}, SAVER \cite{ref45}, HRCN \cite{ref9}, LCDNet+BRL+RR \cite{ref46}.

As can be seen from TABLE \ref{tab4}, in all three datasets, our method is almost the best compared with those view-aware based methods or those metric learning based methods, and our method is also the best compared with the two papers published in 2022 (i.e. VAT  \cite{ref44} and LCDNet+BRL+RR \cite{ref46}).

\subsection{Discussion}

Compared with the traditional supervised contrastive loss, the proposed global-supervised contrastive loss has the advantage that in each iteration, the features of the entire training set can be used, but the disadvantage is that it will occupy more memory, but these extra memory has no relationship with the number of layers of the deep model, so it generally does not affect the running of the program. The proposed global-supervised contrastive loss performs better in datasets with larger training sets, while for smaller training sets, it needs to be combined with traditional supervised contrastive loss.

The advantages of the VABPP are that it is used during testing without affecting the training process, and the view pair distance scaling coefficient matrix  $\Delta^{V\times V}$s provided by us can be used directly, which makes our proposed VABPP method can be easy integrated into other methods. This method also has two shortcomings. One is that it needs to label the orientation information of the training set, and use the orientation information to train an orientation extraction network to extract test set image orientations. Second, since the view-pair distance scaling coefficient matrix  $\Delta^{V\times V}$ is calculated by counting elements of the distance matrix between all training set features, the larger the dataset, the performance will be the better, because the statistical significance of the smaller data set is relatively smaller.

\section{Summary}
In this paper, we propose a global-supervised contrastive loss and a view-aware-based post-processing method to address two challenges in the field of vehicle re-id. The global-supervised contrastive loss has a good effect on the training set with a large number of training IDs. We verify this conclusion by using three different backbones in three datasets widely used in vehicle re-identification. The view-aware-based post-processing method does not affect the training process, because it is only used in testing. We provide the orientation labels of the training set of the three datasets, and also provide the calculated view-pair distance scaling coefficient matrices of the three datasets, which makes it easy to integrate the VABPP method into other methods, and we use experiments to integrate VABPP into the Bag-of-Tricks \cite{ref41} as a baseline commonly used in the field of vehicle re-identification, which verifies the feasibility of this method.

\section*{Acknowledgments}
This work is supported by Shenzhen Key Laboratory of Visual Object Detection and Recognition under Grant No. ZDSYS20190902093015527, National Natural Science Foundation of China under Grant No. 61876051, Science and Technology Plan Project of Guizhou Province, No. Qiankehe Foundation-ZK[2022] General 550.

\section*{Appendix A: The solution process of $\frac{\partial\mathcal{L}_{GSupCon}}{\partial f_i}$ and  $\frac{\partial\mathcal{L}_{LSupCon}}{\partial f_i}$.}
\label{appendix1}

\begin{equation}
	\label{LG_split}
	\begin{array} {ll}
		&\mathcal{L}_{GSupCon}\\
		=&\sum\limits_{i\in I}\frac{-1}{\lvert \widetilde{P}(i) \rvert}\sum\limits_{p\in \widetilde{P}(i)}log\frac{exp(f_i\widetilde{f}_p/\tau)}{\sum\limits_{a\in T}exp(f_i\widetilde{f}_a/\tau)} \\
		=&\frac{-1}{\lvert \widetilde{P}(i) \rvert}\sum\limits_{p\in \widetilde{P}(i)}log\frac{exp(f_i\widetilde{f}_p/\tau)}{\sum\limits_{a\in T}exp(f_i\widetilde{f}_a/\tau)} \\
		&+\sum\limits_{k\in I,k\neq i}\frac{-1}{\lvert \widetilde{P}(k) \rvert}\sum\limits_{p\in \widetilde{P}(k)}log\frac{exp(z_k\widetilde{f}_p/\tau)}{\sum\limits_{a\in T}exp(z_k\widetilde{f}_a/\tau)} \\
		=&\frac{-1}{\lvert\widetilde{P}(i)\rvert}\sum\limits_{p\in\widetilde{P}(i)}\{f_i\widetilde{f}_p/\tau-log[\sum\limits_{a\in T}exp(f_i\widetilde{f}_a/\tau)]\}\\
		&+\sum\limits_{k\in I,k\neq i}\frac{-1}{\lvert \widetilde{P}(k) \rvert}\sum\limits_{p\in \widetilde{P}(k)}log\frac{exp(z_k\widetilde{f}_p/\tau)}{\sum\limits_{a\in T}exp(z_k\widetilde{f}_a/\tau)} \\
		=&\frac{-f_i}{\tau \lvert\widetilde{P}(i)\rvert}\sum\limits_{p\in \widetilde{P}(i)}\widetilde{f}_p-log[\sum\limits_{a\in T}exp(f_i\widetilde{f}_a/\tau)] \\
		&+\sum\limits_{k\in I,k\neq i}\frac{-1}{\lvert \widetilde{P}(k) \rvert}\sum\limits_{p\in \widetilde{P}(k)}log\frac{exp(z_k\widetilde{f}_p/\tau)}{\sum\limits_{a\in T}exp(z_k\widetilde{f}_a/\tau)} \\

	\end{array}
\end{equation}

\begin{small}
	\begin{equation}
		\label{LL_split}
		\begin{array} {ll}
			&\mathcal{L}_{LSupCon}\\
			=&\sum\limits_{i\in I}\frac{-1}{\lvert P(i) \rvert}\sum\limits_{p\in P(i)}log\frac{exp(f_if_p/\tau)}{\sum\limits_{a\in A(i)}exp(f_if_a/\tau)}\\
			=&\frac{-1}{\lvert P(i) \rvert}\sum\limits_{p\in P(i)}log\frac{exp(f_if_p/\tau)}{\sum\limits_{a\in A(i)}exp(f_if_a/\tau)}\\
			&+\sum\limits_{k\in P(i)}\frac{-1}{\lvert P(k) \rvert}\sum\limits_{p\in P(k)}log\frac{exp(z_kf_p/\tau)}{\sum\limits_{a\in A(k)}exp(z_kf_a/\tau)}\\
			&+\sum\limits_{k\in N(i)}\frac{-1}{\lvert P(k) \rvert}\sum\limits_{p\in P(k)}log\frac{exp(z_kf_p/\tau)}{\sum\limits_{a\in A(k)}exp(z_kf_a/\tau)}\\
			=&\frac{-1}{\lvert P(i)\rvert}\sum\limits_{p\in P(i)}\{f_i f_p/\tau-log[\sum\limits_{a\in A(i)}exp(f_i f_a/\tau)]\}\\
			&+\sum\limits_{k\in P(i)}\frac{-1}{\lvert P(k)\rvert}\sum\limits_{p\in P(k)}\{z_k f_p/\tau-log[\sum\limits_{a\in A(k)}exp(z_k f_a/\tau)]\}\\
			&+\sum\limits_{k\in N(i)}\frac{-1}{\lvert P(k)\rvert}\sum\limits_{p\in P(k)}\{z_k f_p/\tau-log[\sum\limits_{a\in A(k)}exp(z_k f_a/\tau)]\}\\
			
			=&log[\sum\limits_{a\in A(i)}exp(f_if_a/\tau)]-\sum\limits_{p\in P(i)}\frac{f_if_p}{\tau\lvert P(i)\rvert}\\
			&+\sum\limits_{k\in P(i)}\{log[\sum\limits_{a\in A(k)}exp(z_kf_a/\tau)] \\
			&-\frac{z_kf_i}{\tau \lvert P(k)\rvert}-\sum\limits_{p\in P(k),p\neq i}\frac{z_kf_p}{\tau\lvert P(k)\rvert}\}\\
			&+\sum\limits_{k\in N(i)}log[\sum\limits_{a\in A(k)}exp(z_kf_a/\tau)]-\sum\limits_{p\in P(k)}\frac{z_kf_p}{\tau\lvert P(k)\rvert}

		\end{array}
	\end{equation}
\end{small}

	\begin{equation}
		\label{LGzi2}
		\begin{array} {ll}
			&\frac{\partial\mathcal{L}_{GSupCon}}{\partial f_i} \\
			=&\frac{-1}{\tau}\{\sum\limits_{p\in\widetilde{P}(i)}\frac{\widetilde{f}_p}{\lvert\widetilde{P}(i)\rvert}-\frac{\sum\limits_{a\in T}\widetilde{f}_a\bigcdot exp(f_i\widetilde{f}_a/\tau)}{\sum\limits_{a\in T}exp(f_i\widetilde{f}_a/\tau)}\} \\
			=&\frac{-1}{\tau}\{\sum\limits_{p\in\widetilde{P}(i)}\frac{\widetilde{f}_p}{\lvert\widetilde{P}(i)\rvert}-\frac{\sum\limits_{p\in \widetilde{P}(i)}\widetilde{f}_p\bigcdot exp(f_i\widetilde{f}_p/\tau)}{\sum\limits_{a\in T}exp(f_i\widetilde{f}_a/\tau)} -\frac{\sum\limits_{n\in \widetilde{N}(i)}\widetilde{f}_n\bigcdot exp(f_i\widetilde{f}_n/\tau)}{\sum\limits_{a\in T}exp(f_i\widetilde{f}_a/\tau)}\} \\
			=&\frac{1}{\tau}\{\sum\limits_{p\in\widetilde{P}(i)}\widetilde{f}_p[\frac{exp(f_i\widetilde{f}_p/\tau)}{\sum\limits_{a\in T}exp(f_i\widetilde{f}_a/\tau)}-\frac{1}{\lvert\widetilde{P}(i)\rvert}] +\frac{\sum\limits_{n\in \widetilde{N}(i)}\widetilde{f}_n\bigcdot exp(f_i\widetilde{f}_n/\tau)}{\sum\limits_{a\in T}exp(f_i\widetilde{f}_a/\tau)}\}

		\end{array}
	\end{equation}

\begin{equation}
	\label{LLzi2}
	\begin{array} {ll}
		&\frac{\partial\mathcal{L}_{LSupCon}}{\partial f_i} \\
		=&\frac{\partial log[\sum\limits_{a\in A(i)}exp(f_if_a/\tau)]}{\partial f_i}-\frac{\partial \{f_i\sum\limits_{p\in P(i)}\frac{f_p}{\tau\lvert P(i)\rvert}\}}{\partial f_i} \\
		&+\sum\limits_{k\in P(i)}\{\frac{\partial log[\sum\limits_{a\in A(k)}exp(z_kf_a/\tau)]}{\partial f_i} - \frac{\partial\frac{z_kf_i}{\tau\lvert P(k)\rvert}}{\partial f_i} \\& - \frac{\partial\sum\limits_{p\in P(k),p\neq i}\frac{z_kf_p}{\tau\lvert P(k)\rvert}}{\partial f_i}
		\} \\
		&+\sum\limits_{k\in N(i)}\{\frac{\partial log[\sum\limits_{a\in A(k)}exp(z_kf_a/\tau)]}{\partial f_i}  - \frac{\partial\sum\limits_{p\in P(k)}\frac{z_kf_p}{\tau\lvert P(k)\rvert}}{\partial f_i}
		\} \\
		=&\frac{\frac{1}{\tau}\sum\limits_{a\in A(i)}f_a\bigcdot exp(f_if_a/\tau)}{\sum\limits_{a\in A(i)}exp(f_if_a/\tau)}-\sum\limits_{p\in P(i)}\frac{f_p}{\tau\lvert P(i)\rvert} \\
		&+\sum\limits_{k\in P(i)}\{\frac{\frac{1}{\tau}z_k\bigcdot exp(z_kf_i/\tau)}{\sum\limits_{a\in A(k)}exp(z_kf_a/\tau)}-\frac{z_k}{\tau\lvert P(k)\rvert}\}\\
		&+\sum\limits_{k\in N(i)}\frac{\frac{1}{\tau}z_k\bigcdot exp(z_kf_i/\tau)}{\sum\limits_{a\in A(k)}exp(z_kf_a/\tau)}\\
		=&\frac{1}{\tau}\{\sum\limits_{p\in P(i)}f_p[\frac{exp(f_if_p/\tau)}{\sum\limits_{a\in A(i)}exp(f_if_a/\tau)}-\frac{1}{\lvert P(i)\rvert}] \\
		&+\sum\limits_{n\in N(i)}\frac{f_n\bigcdot exp(f_if_n/\tau)}{\sum\limits_{a\in A(i)}exp(f_if_a/\tau)}\} \\
		&\frac{1}{\tau}\{\sum\limits_{p\in P(i)}f_p[\frac{exp(f_pf_i/\tau)}{\sum\limits_{a\in A(p)}exp(f_pf_a/\tau)}-\frac{1}{\lvert P(p)\rvert}] \\
		&+\sum\limits_{n\in N(i)}\frac{f_n\bigcdot exp(f_nf_i/\tau)}{\sum\limits_{a\in A(n)}exp(f_nf_a/\tau)}\}

	\end{array}
\end{equation}

\begin{table}[htb]
	\caption{ The universal $\Delta^{V\times V}$ in VeRi-776 dataset. The number of 0,1,2,3,4,5,6 and 7 represent front, rear, left, front left, rear left, right, front right and rear right, respectively.}\label{tab5}
	\centering
	\setlength{\tabcolsep}{1.15mm}
	\begin{tabular}{|c|c|c|c|c|c|c|c|c|}
		\hline
		&0	&1	&2	&3	&4	&5	&6	&7\\
		\hline
		0	&1	&0.8930	&0.7507	&0.9217	&0.8607	&0.6772	&0.9712	&0.8709\\
		\hline
		1	&0.8242	&1	&0.7103	&0.7874	&0.8907	&0.6404	&0.8207	&0.9329\\
		\hline
		2	&0.8641	&0.8858	&1	&0.9112	&0.9222	&0.9600	&0.9369	&0.9021\\
		\hline
		3	&0.8908	&0.8245	&0.7651	&1	&0.8993	&0.6886	&0.8947	&0.8518\\
		\hline
		4	&0.7965	&0.8931	&0.7414	&0.8611	&1	&0.6608	&0.8028	&0.9154\\
		\hline
		5	&1.0297	&1.0550	&1.2682	&1.0834	&1.0857	&1	&1.0763	&1.1352\\
		\hline
		6	&1.0762	&0.9928	&0.9020	&1.0258	&0.9612	&0.7843	&1	&0.9910\\
		\hline
		7	&0.8654	&1.0042	&0.7788	&0.8758	&0.9828	&0.7418	&0.8887	&1\\
		\hline
		
	\end{tabular}
\end{table}

\begin{table}[htb]
	\caption{ The universal $\Delta^{V\times V}$ in VehicleID dataset. The number of 0 and 1 represent front and rear, respectively.}\label{tab6}
	\centering
	\setlength{\tabcolsep}{1.55mm}
	\begin{tabular}{|c|c|c|}
		\hline
		&0&1\\
		\hline
		0&1&0.4597\\
		\hline
		1&0.6455&1\\
		\hline
		
	\end{tabular}
\end{table}

\begin{table}[h]
	\caption{ The universal $\Delta^{V\times V}$ inVERI\_Wild dataset. The number of 0,1,2,3,4 and 5 represent front, rear, rear right, front left, rear left and front right.}\label{tab7}
	\centering
	\setlength{\tabcolsep}{2.55mm}
	\begin{tabular}{|c|c|c|c|c|c|c|}
		\hline
		&0	&1	&2	&3	&4	&5\\
		\hline
		0	&1	&0.6556	&0.6521	&0.8721	&0.6676	&0.9675\\
		\hline
		1	&0.6834	&1	&0.9303	&0.6932	&0.9146	&0.6887\\
		\hline
		2	&0.6658	&0.9112	&1	&0.6982	&0.9687	&0.6972\\
		\hline
		3	&0.8653	&0.6597	&0.6784	&1	&0.7247	&0.9390\\
		\hline
		4	&0.6203	&0.8152	&0.8815	&0.6786	&1	&0.6383\\
		\hline
		5	&0.9478	&0.6473	&0.6690	&0.9272	&0.6731	&1\\
		\hline
		
	\end{tabular}
\end{table}

\section*{Appendix B: The universal $\Delta^{V\times V}$s of the three datasets}
\label{appendix2}

The $\Delta^{V\times V}$s in TABLEs \ref{tab5} and \ref{tab6} are obtained by training with $\mathcal{L}_{GSupCon}$ + $\mathcal{L}_{LSupCon}$, while the $\Delta^{V\times V}$ in TABLE \ref{tab7} is obtained by training with $\mathcal{L}_{GSupCon}$ alone. Take the first row in TABLE \ref{tab5} as an example to understand the meaning of the whole table. Let $q\in Q$ and $g\in G$, where $\mathcal{V}(q)=0$. If $\mathcal{V}(g)=0$, because $\delta(0,0)=1$, so in formula \ref{d_scaled}, we have $d_{scaled}(f_q,f_q|\theta)=d_{unified}(f_q,f_q|\theta)\times 1$. If $\mathcal{V}(g)=1$, because $\delta(0,1)=0.8930$, so in formula \ref{d_scaled}, we have $d_{scaled}(f_q,f_q|\theta)=d_{unified}(f_q,f_q|\theta)\times 0.8930$, and so on. We can see that the larger the value of $\delta(i,j)$, the easier query image with view $i$ to be matched with the images in the gallery set with the view $j$. Generally speaking, the images with the same view is easier to match. We can note that in TABLE \ref{tab5}, we found that when the query image $q$ is with the right view (the penultimate row), all the values are bigger than 1 except $\delta(5,5)=1$, this situation is the opposite of the other rows. The reason is that there are too few images with the right view in the VeRi-776 dataset, and the statistics are not general.


\bibliographystyle{apalike}
\bibliography{refs} 

\begin{thebibliography}{}

\bibitem[Bai et~al., 2018]{ref20}
Bai, Y., Lou, Y., Gao, F., Wang, S., Wu, Y., and Duan, L.-Y. (2018).
\newblock Group-sensitive triplet embedding for vehicle reidentification.
\newblock {\em IEEE Transactions on Multimedia}, 20(9):2385--2399.

\bibitem[Banerjee et~al., 2021]{ref3}
Banerjee, R., De, S., and Dey, S. (2021).
\newblock A survey on various deep learning algorithms for an efficient facial
  expression recognition system.
\newblock {\em International Journal of Image and Graphics}, page 2240005.

\bibitem[Chen et~al., 2020a]{ref11}
Chen, T., Kornblith, S., Norouzi, M., and Hinton, G. (2020a).
\newblock A simple framework for contrastive learning of visual
  representations.
\newblock In {\em International conference on machine learning}, pages
  1597--1607. PMLR.

\bibitem[Chen et~al., 2020b]{ref35}
Chen, T.-S., Liu, C.-T., Wu, C.-W., and Chien, S.-Y. (2020b).
\newblock Orientation-aware vehicle re-identification with semantics-guided
  part attention network.
\newblock In {\em European Conference on Computer Vision}, pages 330--346.
  Springer.

\bibitem[Chu et~al., 2019]{ref15}
Chu, R., Sun, Y., Li, Y., Liu, Z., Zhang, C., and Wei, Y. (2019).
\newblock Vehicle re-identification with viewpoint-aware metric learning.
\newblock In {\em Proceedings of the IEEE/CVF International Conference on
  Computer Vision}, pages 8282--8291.

\bibitem[Fu et~al., 2022]{ref46}
Fu, X., Peng, J., Jiang, G., and Wang, H. (2022).
\newblock Learning latent features with local channel drop network for vehicle
  re-identification.
\newblock {\em Engineering Applications of Artificial Intelligence},
  107:104540.

\bibitem[Ge et~al., 2015]{ref39}
Ge, R., Huang, F., Jin, C., and Yuan, Y. (2015).
\newblock Escaping from saddle points-online stochastic gradient for tensor
  decomposition.
\newblock In {\em Conference on learning theory}, pages 797--842. PMLR.

\bibitem[Ghosh et~al., 2021]{ref23}
Ghosh, A., Shanmugalingam, K., and Lin, W.-Y. (2021).
\newblock Relation preserving triplet mining for stabilizing the triplet loss
  in vehicle re-identification.
\newblock {\em arXiv preprint arXiv:2110.07933}.

\bibitem[He et~al., 2020]{ref12}
He, K., Fan, H., Wu, Y., Xie, S., and Girshick, R. (2020).
\newblock Momentum contrast for unsupervised visual representation learning.
\newblock In {\em Proceedings of the IEEE/CVF conference on computer vision and
  pattern recognition}, pages 9729--9738.

\bibitem[Hoffer and Ailon, 2015]{ref7}
Hoffer, E. and Ailon, N. (2015).
\newblock Deep metric learning using triplet network.
\newblock In {\em International workshop on similarity-based pattern
  recognition}, pages 84--92. Springer.

\bibitem[Huang et~al., 2020]{ref2}
Huang, C., Peng, Z., Xu, Y., Chen, F., Jiang, Q., Zhang, Y., Jiang, G., and Ho,
  Y.-S. (2020).
\newblock Online learning-based multi-stage complexity control for live video
  coding.
\newblock {\em IEEE Transactions on Image Processing}, 30:641--656.

\bibitem[Huynh, 2021]{ref24}
Huynh, S.~V. (2021).
\newblock A strong baseline for vehicle re-identification.
\newblock In {\em Proceedings of the IEEE/CVF Conference on Computer Vision and
  Pattern Recognition}, pages 4147--4154.

\bibitem[Jin et~al., 2021]{ref37}
Jin, Y., Li, C., Li, Y., Peng, P., and Giannopoulos, G.~A. (2021).
\newblock Model latent views with multi-center metric learning for vehicle
  re-identification.
\newblock {\em IEEE Transactions on Intelligent Transportation Systems},
  22(3):1919--1931.

\bibitem[Khorramshahi et~al., 2019]{ref13}
Khorramshahi, P., Kumar, A., Peri, N., Rambhatla, S.~S., Chen, J.-C., and
  Chellappa, R. (2019).
\newblock A dual-path model with adaptive attention for vehicle
  re-identification.
\newblock In {\em Proceedings of the IEEE/CVF International Conference on
  Computer Vision}, pages 6132--6141.

\bibitem[Khorramshahi et~al., 2020]{ref45}
Khorramshahi, P., Peri, N., Chen, J.-c., and Chellappa, R. (2020).
\newblock The devil is in the details: Self-supervised attention for vehicle
  re-identification.
\newblock In {\em European Conference on Computer Vision}, pages 369--386.
  Springer.

\bibitem[Khosla et~al., 2020]{ref10}
Khosla, P., Teterwak, P., Wang, C., Sarna, A., Tian, Y., Isola, P., Maschinot,
  A., Liu, C., and Krishnan, D. (2020).
\newblock Supervised contrastive learning.
\newblock In Larochelle, H., Ranzato, M., Hadsell, R., Balcan, M., and Lin, H.,
  editors, {\em Advances in Neural Information Processing Systems}, volume~33,
  pages 18661--18673. Curran Associates, Inc.

\bibitem[Kuma et~al., 2019]{ref21}
Kuma, R., Weill, E., Aghdasi, F., and Sriram, P. (2019).
\newblock Vehicle re-identification: an efficient baseline using triplet
  embedding.
\newblock In {\em 2019 International Joint Conference on Neural Networks
  (IJCNN)}, pages 1--9. IEEE.

\bibitem[Li et~al., 2020]{ref36}
Li, Y., Liu, K., Jin, Y., Wang, T., and Lin, W. (2020).
\newblock Varid: Viewpoint-aware re-identification of vehicle based on triplet
  loss.
\newblock {\em IEEE Transactions on Intelligent Transportation Systems}.

\bibitem[Liu et~al., 2016]{ref5}
Liu, X., Liu, W., Ma, H., and Fu, H. (2016).
\newblock Large-scale vehicle re-identification in urban surveillance videos.
\newblock In {\em 2016 IEEE international conference on multimedia and expo
  (ICME)}, pages 1--6. IEEE.

\bibitem[Liu et~al., 2020]{ref42}
Liu, X., Liu, W., Zheng, J., Yan, C., and Mei, T. (2020).
\newblock Beyond the parts: Learning multi-view cross-part correlation for
  vehicle re-identification.
\newblock In {\em Proceedings of the 28th ACM International Conference on
  Multimedia}, pages 907--915.

\bibitem[Lou et~al., 2019a]{ref22}
Lou, Y., Bai, Y., Liu, J., Wang, S., and Duan, L. (2019a).
\newblock Veri-wild: A large dataset and a new method for vehicle
  re-identification in the wild.
\newblock In {\em Proceedings of the IEEE/CVF conference on computer vision and
  pattern recognition}, pages 3235--3243.

\bibitem[Lou et~al., 2019b]{ref30}
Lou, Y., Bai, Y., Liu, J., Wang, S., and Duan, L.-Y. (2019b).
\newblock Embedding adversarial learning for vehicle re-identification.
\newblock {\em IEEE Transactions on Image Processing}, 28(8):3794--3807.

\bibitem[Luo et~al., 2019]{ref41}
Luo, H., Gu, Y., Liao, X., Lai, S., and Jiang, W. (2019).
\newblock Bag of tricks and a strong baseline for deep person
  re-identification.
\newblock In {\em Proceedings of the IEEE/CVF conference on computer vision and
  pattern recognition workshops}, pages 0--0.

\bibitem[Masters and Luschi, 2018]{ref38}
Masters, D. and Luschi, C. (2018).
\newblock Revisiting small batch training for deep neural networks.
\newblock {\em arXiv preprint arXiv:1804.07612}.

\bibitem[Pan et~al., 2020]{ref28}
Pan, M., Zhu, X., Li, Y., Qian, J., and Liu, P. (2020).
\newblock Mrnet: A keypoint guided multi-scale reasoning network for vehicle
  re-identification.
\newblock In {\em International Conference on Neural Information Processing},
  pages 469--478. Springer.

\bibitem[Schroff et~al., 2015]{ref6}
Schroff, F., Kalenichenko, D., and Philbin, J. (2015).
\newblock Facenet: A unified embedding for face recognition and clustering.
\newblock In {\em Proceedings of the IEEE conference on computer vision and
  pattern recognition}, pages 815--823.

\bibitem[Sharma et~al., 2021]{ref4}
Sharma, S., Gupta, S., Kumar, N., and Arora, T. (2021).
\newblock Postal automation system in gurmukhi script using deep learning.
\newblock {\em International Journal of Image and Graphics}, page 2350005.

\bibitem[Shen et~al., 2017]{ref17}
Shen, Y., Xiao, T., Li, H., Yi, S., and Wang, X. (2017).
\newblock Learning deep neural networks for vehicle re-id with
  visual-spatio-temporal path proposals.
\newblock In {\em Proceedings of the IEEE International Conference on Computer
  Vision}, pages 1900--1909.

\bibitem[Sun et~al., 2020]{ref16}
Sun, Z., Nie, X., Xi, X., and Yin, Y. (2020).
\newblock Cfvmnet: A multi-branch network for vehicle re-identification based
  on common field of view.
\newblock In {\em Proceedings of the 28th ACM International Conference on
  Multimedia}, pages 3523--3531.

\bibitem[Teng et~al., 2020]{ref33}
Teng, S., Zhang, S., Huang, Q., and Sebe, N. (2020).
\newblock Multi-view spatial attention embedding for vehicle re-identification.
\newblock {\em IEEE Transactions on Circuits and Systems for Video Technology},
  31(2):816--827.

\bibitem[Teng et~al., 2021]{ref43}
Teng, S., Zhang, S., Huang, Q., and Sebe, N. (2021).
\newblock Viewpoint and scale consistency reinforcement for uav vehicle
  re-identification.
\newblock {\em International Journal of Computer Vision}, 129(3):719--735.

\bibitem[Wang et~al., 2021]{ref32}
Wang, Q., Min, W., Han, Q., Yang, Z., Xiong, X., Zhu, M., and Zhao, H. (2021).
\newblock Viewpoint adaptation learning with cross-view distance metric for
  robust vehicle re-identification.
\newblock {\em Information Sciences}, 564:71--84.

\bibitem[Wang et~al., 2017]{ref25}
Wang, Z., Tang, L., Liu, X., Yao, Z., Yi, S., Shao, J., Yan, J., Wang, S., Li,
  H., and Wang, X. (2017).
\newblock Orientation invariant feature embedding and spatial temporal
  regularization for vehicle re-identification.
\newblock In {\em Proceedings of the IEEE international conference on computer
  vision}, pages 379--387.

\bibitem[Yu et~al., 2021]{ref40}
Yu, W., Hu, B., Hu, Y., Lan, T., You, Y., and Yin, D. (2021).
\newblock Revisiting the loss weight adjustment in object detection.
\newblock {\em arXiv preprint arXiv:2103.09488}.

\bibitem[Yu et~al., 2022]{ref44}
Yu, Z., Pei, J., Zhu, M., Zhang, J., and Li, J. (2022).
\newblock Multi-attribute adaptive aggregation transformer for vehicle
  re-identification.
\newblock {\em Information Processing \& Management}, 59(2):102868.

\bibitem[Zhao et~al., 2021a]{ref8}
Zhao, J., Qi, F., Ren, G., and Xu, L. (2021a).
\newblock Phd learning: Learning with pompeiu-hausdorff distances for
  video-based vehicle re-identification.
\newblock In {\em Proceedings of the IEEE/CVF Conference on Computer Vision and
  Pattern Recognition}, pages 2225--2235.

\bibitem[Zhao et~al., 2021b]{ref9}
Zhao, J., Zhao, Y., Li, J., Yan, K., and Tian, Y. (2021b).
\newblock Heterogeneous relational complement for vehicle re-identification.
\newblock In {\em Proceedings of the IEEE/CVF International Conference on
  Computer Vision}, pages 205--214.

\bibitem[Zheng et~al., 2021]{ref14}
Zheng, B., Lei, Z., Tang, C., Wang, J., Liao, Z., Yu, Z., and Xie, Y. (2021).
\newblock Oerff: A vehicle re-identification method based on orientation
  estimation and regional feature fusion.
\newblock {\em IEEE Access}, 9:66661--66674.

\bibitem[Zhou et~al., 2018]{ref26}
Zhou, Y., Liu, L., and Shao, L. (2018).
\newblock Vehicle re-identification by deep hidden multi-view inference.
\newblock {\em IEEE Transactions on Image Processing}, 27(7):3275--3287.

\bibitem[Zhou and Shao, 2017]{ref29}
Zhou, Y. and Shao, L. (2017).
\newblock Cross-view gan based vehicle generation for re-identification.
\newblock In {\em BMVC}, volume~1, pages 1--12.

\bibitem[Zhou and Shao, 2018]{ref27}
Zhou, Y. and Shao, L. (2018).
\newblock Vehicle re-identification by adversarial bi-directional lstm network.
\newblock In {\em 2018 IEEE Winter Conference on Applications of Computer
  Vision (WACV)}, pages 653--662. IEEE.

\bibitem[Zhu et~al., 2018a]{ref18}
Zhu, J., Zeng, H., Du, Y., Lei, Z., Zheng, L., and Cai, C. (2018a).
\newblock Joint feature and similarity deep learning for vehicle
  re-identification.
\newblock {\em IEEE Access}, 6:43724--43731.

\bibitem[Zhu et~al., 2018b]{ref19}
Zhu, J., Zeng, H., Lei, Z., Liao, S., Zheng, L., and Cai, C. (2018b).
\newblock A shortly and densely connected convolutional neural network for
  vehicle re-identification.
\newblock In {\em 2018 24th International Conference on Pattern Recognition
  (ICPR)}, pages 3285--3290. IEEE.

\bibitem[Zhu et~al., 2020]{ref34}
Zhu, X., Luo, Z., Fu, P., and Ji, X. (2020).
\newblock Voc-reid: Vehicle re-identification based on
  vehicle-orientation-camera.
\newblock In {\em Proceedings of the IEEE/CVF Conference on Computer Vision and
  Pattern Recognition Workshops}, pages 602--603.

\end{thebibliography}

\end{document}